\documentclass{article} %
\usepackage{iclr2026_conference,times}

    \PassOptionsToPackage{authoryear, sort&compress, round}{natbib}

\usepackage{amsmath,amsfonts,bm}

\def\eqref#1{equation~\ref{#1}}

\def\1{\bm{1}}

\DeclareMathAlphabet{\mathsfit}{\encodingdefault}{\sfdefault}{m}{sl}
\SetMathAlphabet{\mathsfit}{bold}{\encodingdefault}{\sfdefault}{bx}{n}

\DeclareMathOperator*{\argmax}{arg\,max}

\iclrfinalcopy %

\usepackage[utf8]{inputenc} %
\usepackage[T1]{fontenc}    %
\usepackage{hyperref}
\usepackage{url}
\usepackage{algorithm}
\usepackage{algorithmic}
\usepackage{bbm}
\usepackage{CJKutf8}
\usepackage{graphicx}
\usepackage{subfigure}
\usepackage{booktabs}
\usepackage{multirow}
\usepackage{dsfont}
\usepackage{stackengine}
\usepackage{makecell}
\usepackage{wrapfig}
\usepackage[table,xcdraw,svgnames,HTML]{xcolor}
\usepackage{enumitem}

\usepackage{amsmath}
\usepackage{amssymb}
\usepackage{mathtools}
\usepackage{amsthm}
\usepackage{dashrule}
\usepackage{titletoc}
\usepackage{lipsum}
\usepackage{caption}
\usepackage[most]{tcolorbox}
\definecolor{mydarkgreen}{RGB}{0,100,0}
\tcbuselibrary{breakable}
\newtcolorbox{mycustombox}[1]{
  enhanced,
  colback=black!5!white,
  colframe=black!75!white,
  boxrule=0.4pt,
  coltitle=white,
  title=#1,
  titlerule=0.4pt,
  fontupper=\small,
  fonttitle=\small,
  before upper={\par\smallskipamount},
  breakable,
  left=3mm,
  right=3mm,
  top=2mm,
  bottom=2mm,
  boxsep=1mm,
}

\usepackage[capitalize]{cleveref}

\definecolor{mediumpersianblue}{rgb}{0.0, 0.4, 0.65}
\definecolor{citecolor}{RGB}{0, 80, 200}
\definecolor{linkcolor}{RGB}{200, 80, 0}
\hypersetup{colorlinks,linkcolor={linkcolor},citecolor={citecolor},urlcolor={citecolor}} 

\newcommand{\Method}{Wide-In, Narrow-Out}

\newcommand{\method}{WINO}

\Crefformat{equation}{#2Eq.~(#1)#3}

\Crefformat{figure}{#2Fig.~#1#3}
\Crefformat{table}{#2Tab.~#1#3}
\Crefformat{assumption}{#2Assumption~#1#3}
\Crefformat{theorem}{#2Theorem~#1#3}
\Crefname{assumption}{Assumption}{Assumptions}

\Crefformat{section}{#2Sec.~#1#3}

\Crefformat{appendix}{#2Appendix~#1#3}
\Crefformat{algorithm}{#2Algorithm~#1#3}

\newcommand{\titlename}{Wide-In, Narrow-Out: Revokable Decoding for Efficient and Effective DLLMs}

\theoremstyle{plain}

\theoremstyle{definition}

\theoremstyle{remark}

\usepackage{nicematrix}

\colorlet{colorours}{red!7}

\usepackage{tikz}
\usetikzlibrary{tikzmark}
\makeatletter
\newcommand*\myfontsize{%
  \@setfontsize\myfontsize{7}{8}%
}
\makeatother
\newcommand{\mytextbox}[2]{\tikzmarknode[draw=#1,thick,inner sep=2pt]{test}{\myfontsize #2}}

\newcommand{\specialtoken}[2]{\mytextbox{#2}{\text{\textcolor{#2}{#1}}}}

\title{\titlename}

\author{%
Feng Hong$^{1,*}$ \,  Geng Yu$^{1,*}$ \, Yushi Ye$^1$ \, Haicheng Huang$^1$ \\
\textbf{Huangjie Zheng}$^2$ \,
\textbf{Ya Zhang}$^{3}$ \, \textbf{Yanfeng Wang}$^{3}$ \, \textbf{Jiangchao Yao}$^{1}$ \\
$^{1}$Cooperative Medianet Innovation Center, Shanghai Jiao Tong University \\ 
$^2$Apple \, $^{3}$School of Artificial Intelligence, Shanghai Jiao Tong University \\
\texttt{\{feng.hong, Sunarker\}@sjtu.edu.cn}
}

\begin{document}

\maketitle
\begingroup
\renewcommand\thefootnote{}\footnotetext{$^*$ Equal contribution}
\endgroup

\begin{abstract}
    Diffusion Large Language Models (DLLMs) have emerged as a compelling alternative to Autoregressive models, designed for fast parallel generation. However, existing DLLMs are plagued by a severe quality-speed trade-off, where faster parallel decoding leads to significant performance degradation. We attribute this to the irreversibility of standard decoding in DLLMs, which is easily polarized into the wrong decoding direction along with early error context accumulation. To resolve this, we introduce {\Method} ({\method}), a training-free decoding algorithm that enables \emph{revokable decoding} in DLLMs. {\method} employs a parallel draft-and-verify mechanism, aggressively drafting multiple tokens while simultaneously using the model's bidirectional context to verify and re-mask suspicious ones for refinement. Verified in open-source DLLMs like LLaDA and MMaDA, {\method} is shown to decisively improve the quality-speed trade-off. For instance, on the GSM8K math benchmark, it accelerates inference by 6$\times$ while improving accuracy by 2.58\%; on Flickr30K captioning, it achieves a 10$\times$ speedup with higher performance. More comprehensive experiments are conducted to demonstrate the superiority and provide an in-depth understanding of {\method}. 
    \footnote{\ Code: \href{https://github.com/Feng-Hong/WINO-DLLM}{https://github.com/Feng-Hong/WINO-DLLM}}
\end{abstract}
    
\section{Introduction}

Autoregressive (AR) large language models~\citep{radford2018improving, radford2019language}, such as the GPT series~\citep{openai2022chatgpt}, have shown impressive performance in a ranging of language tasks. However, their foundational token-by-token generation mechanism introduces inherent limitations, including severe inference latency, susceptibility to error propagation~\citep{DBLP:journals/corr/abs-2310-12397,DBLP:journals/corr/abs-2310-08118}, and challenges in maintaining global coherence~\citep{mei2025surveycontextengineeringlarge}. In response, Diffusion Large Language Models (DLLMs) have emerged as a compelling non-autoregressive alternative, architected to overcome these bottlenecks. By generating tokens simultaneously~\citep{DBLP:conf/nips/LiTGLH22}, DLLMs theoretically enable massive inference acceleration, while their native bidirectional attention offers improved consistency. The immense potential of DLLMs has been showcased by proprietary, closed-source systems (\emph{e.g.}, Mercury Coder~\citep{inceptionlabs2025mercury} and Gemini Diffusion~\citep{deepmind2025geminidiffusion}), which have demonstrated astonishing speeds exceeding 1,000 tokens per second, serving as a powerful proof-of-concept.

\begin{figure*}[t]
\centering
\includegraphics[width=0.92\textwidth]{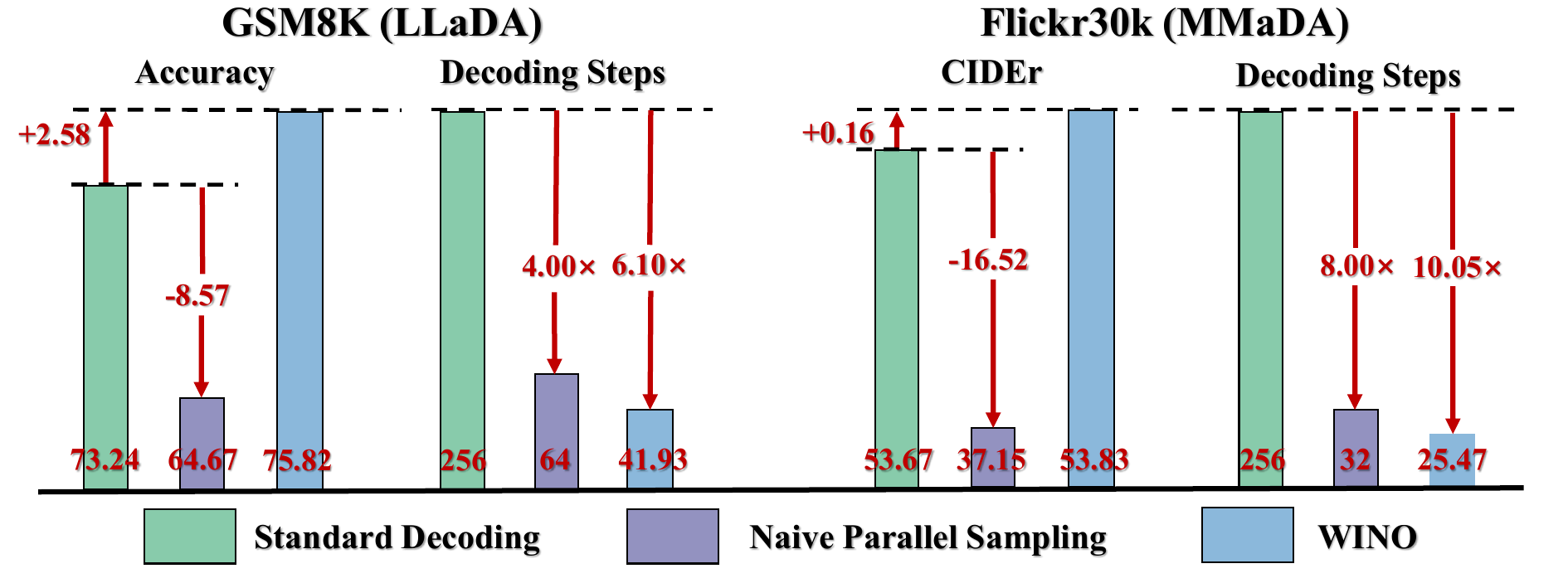}
\vspace{-10pt}
\caption{Demonstration of speedup and performance improvement of {\method} over standard decoding and naive parallel sampling evaluated on GSM8K with LLaDA and Flickr30K with MMaDA. The standard decoding unmasks 1 token per decoding step, while the naive parallel sampling unmasks $M(>1)$ tokens per decoding step. We set $M=4$ for GSM8K and $M=8$ for Flickr30K.}
\label{fig: speedup_demo}
\vspace{-10pt}
\end{figure*}

Despite this promise, the performance of open-source DLLMs has been still disappointing. One critical bottleneck is that they are caught in a severe quality-speed trade-off dilemma. Specifically, to achieve high-quality output, these models are often forced to decode slowly, generating just one token at a time, which negates their primary architectural advantage. As shown in \cref{fig: speedup_demo}, attempting to accelerate inference by generating multiple tokens in parallel invariably leads to a significant degradation in output quality~\citep{DBLP:journals/corr/abs-2502-09992}. This stark trade-off has largely prevented the open-source DLLMs from becoming a viable, high-performance alternative to their AR counterparts.

We attribute this trade-off to a fundamental flaw in DLLMs~\citep{DBLP:conf/nips/SahooASGMCRK24, DBLP:conf/iclr/OuNXZSLL25}: its irreversibility of the standard decoding process. Specifically, the standard generation in diffusion steps typically begins with a sequence of \specialtoken{[MASK]}{gray} tokens, which are then filled in greedily. Once a token is decoded, the decision is final and cannot be revised, even as more informative context becomes available in later steps. However, this is challenging for parallel decoding, where initial tokens are generated with very limited information, easily causing early errors to become permanently embedded, accumulated and propagated throughout the output. Therefore, such a rigid process essentially prevents DLLMs from using their greatest strength of bidirectional attention~\citep{seo2016bidirectional} to refine the early errors when the context progressively becomes rich.

To resolve this problem, we introduce {\Method} ({\method}), a novel decoding algorithm that enables revokable decoding for DLLMs. {\method} employs a novel draft-and-verify procedure that operates in parallel. At each step, a draft module aggressively proposes multiple new tokens based on a lenient threshold (the ``Wide-In''). Concurrently, a verify module leverages the newly enriched global context to re-evaluate all previously generated tokens. Any token that fails a stricter verification check is re-masked for refinement in a future step (the ``Narrow-Out''). This mechanism brings two merits: 1) it breaks the irreversibility of the conventional decoding in DLLMs, allowing the early error to be corrected for better performance; 2) it permits more aggressive token generation in each diffusion step for faster speedup with quality guarantee. Besides, our {\method} is training-free and play-and-plug, which enables the general DLLMs to be both highly efficient and effective.

Our extensive experiments show that when applied to existing open-source models like LLaDA~\citep{DBLP:journals/corr/abs-2502-09992} and MMaDA~\citep{DBLP:journals/corr/abs-2505-15809}, {\method} achieves massive speedups, and also consistently improves model accuracy on both language and visual-language tasks. For instance, as shown in \cref{fig: speedup_demo}, on the GSM8K~\citep{DBLP:journals/corr/abs-2110-14168} math reasoning benchmark, {\method} accelerates inference by 6$\times$ while simultaneously increasing accuracy by 2.58\%, and on Flickr30K~\citep{DBLP:journals/tacl/YoungLHH14} image captioning benchmark, it speedup decoding by 10$\times$ with even higher performance. By making decoding revokable, {\method}  unlocks the latent power of DLLMs in this area.

\section{Related Work}
\textbf{Diffusion-based Language Models.}
Diffusion models~\citep{DBLP:journals/corr/Sohl-DicksteinW15,DBLP:conf/nips/HoJA20,DBLP:conf/iclr/0011SKKEP21}, originally popularized in image generation~\citep{DBLP:conf/cvpr/RombachBLEO22,DBLP:conf/icml/NicholDRSMMSC22,DBLP:conf/nips/SahariaCSLWDGLA22}, have recently gained attention as an alternative to autoregressive language models (ARLMs) for text generation. This expansion from continuous domain to discrete domain is first studied by~\cite{DBLP:journals/corr/Sohl-DicksteinW15}. Subsequently, D3PM~\citep{DBLP:conf/nips/AustinJHTB21} provides a general framework which models the diffusion forward process as a discrete state Markov chain defined by the multiplication of specific transition matrices over discrete time steps. ~\cite{DBLP:conf/nips/CampbellBBRDD22} later expands D3PM to a continuous time setting, utilizing the theory of continuous time Markov chain(CTMC).  
More recently, research on masked diffusion models(MDMs)
~\citep{DBLP:conf/nips/ShiHWDT24} derived from the absorbing state diffusion in D3PM has shown promising results both in small-scale models (\emph{e.g.}, MDLM~\citep{DBLP:conf/nips/SahooASGMCRK24} and RADD~\citep{DBLP:conf/iclr/OuNXZSLL25}) and large-scale implementations (\emph{e.g.}, LLaDA~\citep{DBLP:journals/corr/abs-2502-09992} and Dream~\citep{dream2025}). Extending this line of work, MMaDA~\citep{DBLP:journals/corr/abs-2505-15809} introduces a novel class of multimodal large diffusion models featuring a shared probabilistic formulation and a modality-agnostic architecture. 

\textbf{DLLM Acceleration Techniques.} The existing acceleration study for DLLMs falls into two directions: KV cache and sampling compression. The former targets to build the KV cache for DLLMs due to its bidirectional full attention mechanism, unlike the causal attention of ARLMs. Typical works like Block Diffusion~\citep{DBLP:conf/iclr/ArriolaGCYQHSK25}, Fast-dLLM-cache~\citep{DBLP:journals/corr/abs-2505-22618} and dLLM-cache~\citep{DBLP:journals/corr/abs-2506-06295} respectively explore different caching mechanisms, which shows promising performance for speedup. Note that this direction is out of the scope of our work here. The latter direction focuses on optimizing the sampling process itself. For the classic low-confidence remasking strategy, several works have introduced novel sampling strategies to dynamically adjust the number of tokens predicted in parallel, thereby improving inference efficiency. Fast-dLLM-parallel~\citep{DBLP:journals/corr/abs-2505-22618} adopts a straightforward approach by selecting tokens with confidence scores exceeding a predefined threshold. Meanwhile, ~\cite{DBLP:journals/corr/abs-2505-24857} propose an entropy-bounded (EB) sampler, a drop-in replacement for conventional samplers that leverages an entropy-based unmasking procedure to dynamically decode multiple tokens per step while maintaining a predefined error tolerance. Although our {\method} brings the acceleration promise due to sampling compression, different from these works, we explore to address the inherent limitation of standard decoding in DLLMs.

\section{Preliminary: Decoding Process for DLLMs}
Given a prompt $X$, a DLLM is designed to generate a response $Y = [y_1, y_2, \ldots, y_L]$ with a pre-defined response length $L$. The response sequence is initialized as all special mask tokens, $Y^{(0)} = [\specialtoken{[MASK]}{gray},\specialtoken{[MASK]}{gray},\ldots,\specialtoken{[MASK]}{gray}]$. The decoding process iteratively refines the response sequence $Y^{(k)}$ over a total of $K$ denoising steps. In the following, we detail the case of $K = L$ (\emph{i.e.}, decoding one token per step), as existing models typically achieve optimal performance under this setting~\citep{DBLP:journals/corr/abs-2502-09992}. 

At step $k$, the goal of decoding is to refine the sequence $Y^{(k-1)}$ into $Y^{(k)}$. Given the token vocabulary $V$ and the model parameterized with $\theta$, the model estimates the probability distribution over the response sequence as $p_\theta(\hat{Y}|X, Y^{(k-1)})$. As a common example, in high-confidence greedy decoding, $Y^{(k)}$ is obtained by unmasking the most confident \specialtoken{[MASK]}{gray} token based on $Y^{(k-1)}$, \emph{i.e.},

\begin{equation}
\begin{aligned}
l^{(k)} &= \mathop{\arg\max}_{l\in\{l|y^{(k-1)}_l=\specialtoken{[MASK]}{gray}\}} \left( \max_{v \in V} p_\theta(\hat{y}_l = v | X, Y^{(k-1)}) \right),\\
y^{(k)}_l &= 
\begin{cases}
\argmax\limits_{v \in V} p_\theta(\hat{y}_l = v | X, Y^{(k-1)}), & \text{if } l = l^{(k)}, \\
y^{(k-1)}_l, & \text{otherwise},
\end{cases}
\quad \forall l \in \{1, 2, \dots, L\}.
\end{aligned}
\end{equation}

After completing all $K$ decoding steps, the final generated response is $Y = Y^{(K)}$. Existing DLLMs, such as LLaDA~\citep{DBLP:journals/corr/abs-2502-09992} and MMaDA~\citep{DBLP:journals/corr/abs-2505-15809}, can also accelerate the decoding process via naive parallel sampling by generating multiple tokens (\emph{e.g.}, 2 or 4) per step. However, empirical results reveal that such strategies often result in substantial performance degradation, limiting their practical effectiveness despite the computational speedup~\citep{DBLP:journals/corr/abs-2502-09992}.

\textbf{Semi-Autoregressive Diffusion Decoding.} This strategy is widely adopted by DLLMs like LLaDA~\citep{DBLP:journals/corr/abs-2502-09992} and MMaDA~\citep{DBLP:journals/corr/abs-2505-15809}, which involves splitting the response sequence into multiple blocks and decoding them sequentially from left to right. Within each block, the typical diffusion decoding strategy described above is applied.

\section{Method}
\subsection{Key Limitation of decoding process for DLLMs}
While architecturally suited for parallelism, DLLMs face a critical bottleneck that hinders effective multi-token decoding. During the early generation stages, the sparse context often causes the model to produce low-quality or contradictory tokens when decoding in parallel. 
This issue is significantly exacerbated by the standard decoding process due to its irreversibility nature.
As these flawed initial predictions are permanently locked in, they inevitably propagate and degrade the final generation quality in the progressive diffusion steps, forcing an undesirable trade-off between the speed of parallel decoding and the quality of serial decoding.

To improve the trade-off, one critical point is to abandon the assumption of irreversibility in the standard decoding of DLLMs, and build a process of \emph{revokable decoding} for progressive refinement. This principle empowers the model to iteratively refine its initial parallel outputs. As more context emerges during generation, the model can correct its preliminary predictions. Such a mechanism effectively addresses the core conflict by marrying the efficiency of parallel generation with the accuracy of context-driven corrections.

\begin{figure*}[t]
    \centering
    \subfigure[An overview of {\method}.]{
    \begin{minipage}{0.62\textwidth}
    \centering     
    \includegraphics[width=\textwidth]{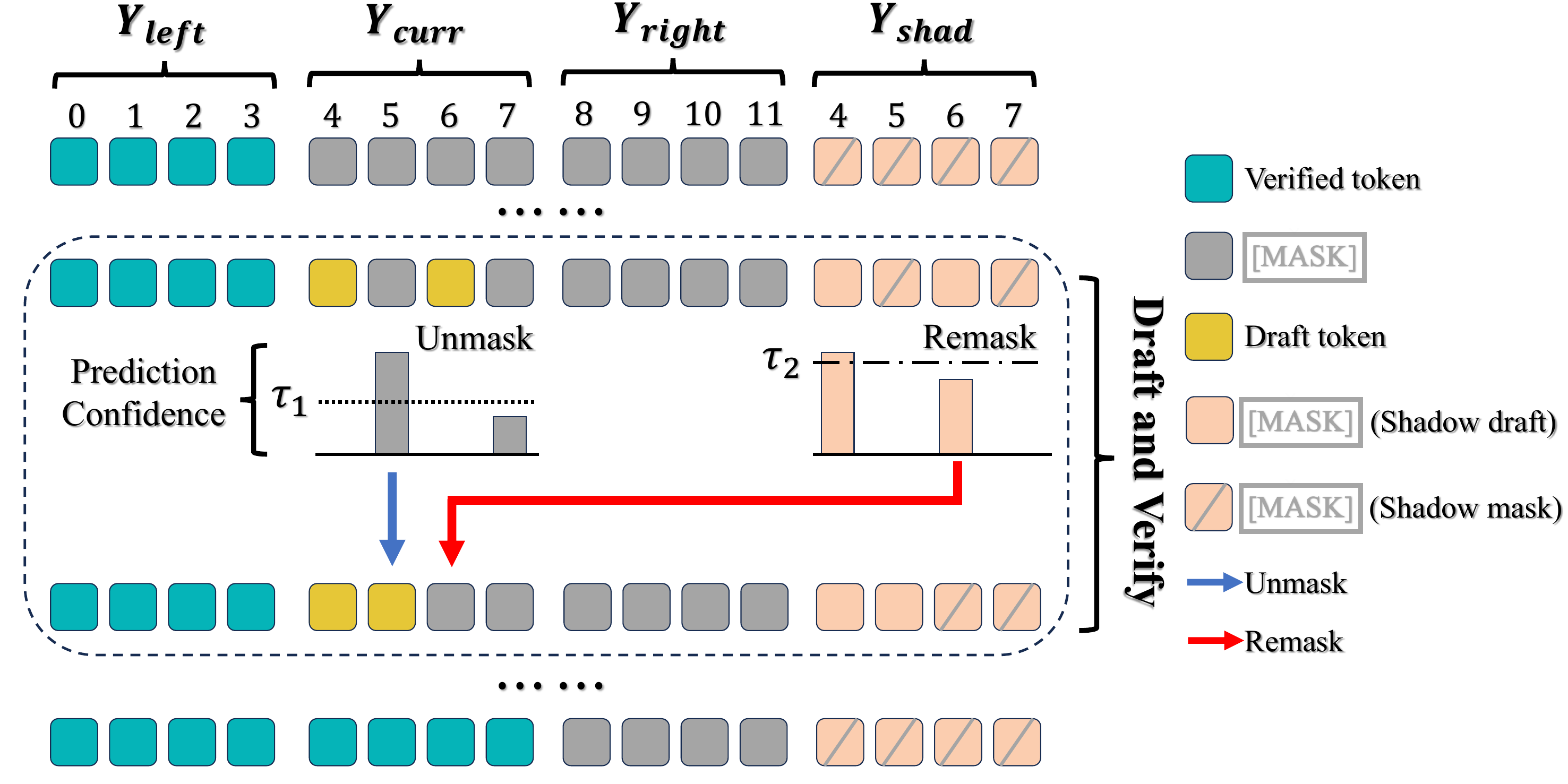}
    \label{fig:method}
    \end{minipage}
    }
\subfigure[Attention Mask.]{
    \begin{minipage}{0.31\textwidth}
    \centering     
    \includegraphics[width=\textwidth]{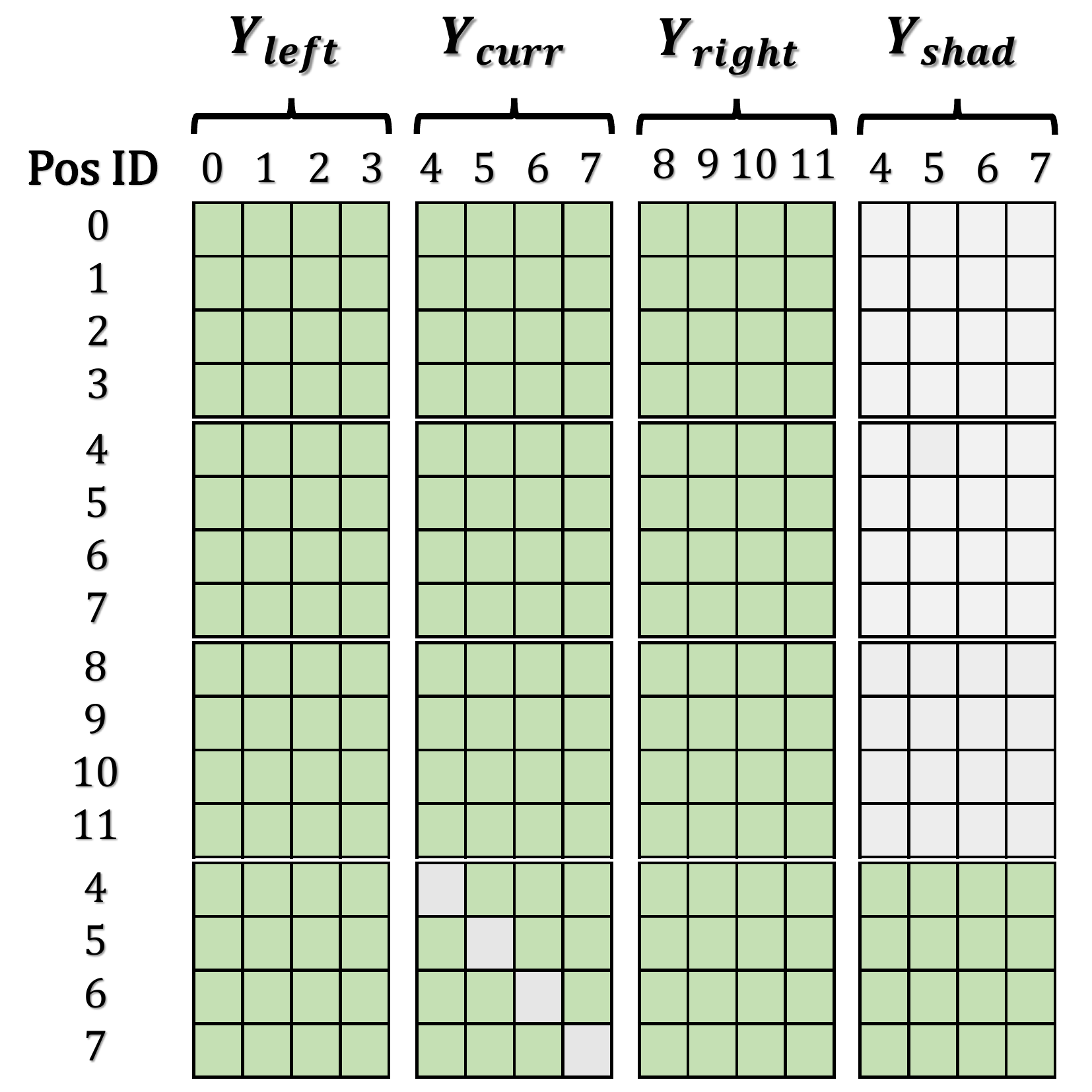}
    \label{fig:attn_mask}
    \end{minipage}
    }
\vspace{-10pt}
\caption{(a) An overview of {\method}. (b) Illustration of our designed attention mask. The green squares denote 1, the grey squares denote 0, and ``Pos ID'' is short for position ID. Verified tokens refer to tokens in the prompt $X$ or previously decoded blocks. Draft tokens denote tokens in the current block that are unmasked up to the current decoding step. \specialtoken{[MASK]}{gray} (shadow draft) refer to tokens in the shadow block whose position IDs correspond to the draft tokens while \specialtoken{[MASK]}{gray} (shadow mask) refer to the remaining tokens in the shadow block.} 
\vspace{-12pt}
\end{figure*}

\subsection{Iterative Refinement via Parallel Draft-and-Verify}
Motivated by the above analysis and the design intuition, we propose a parallel Draft-and-Verify framework to enable revokable decoding for more efficient and higher-quality generation in DLLMs.

Specifically, our framework performs two modules in parallel at each decoding step: 1) Draft: aggressively unmasks multiple \specialtoken{[MASK]}{gray} tokens into candidate meaningful tokens; 2)	Verify: evaluates all currently unmasked tokens and re-masks those deemed low-quality for further refinement. We adopt the most common and general semi-autoregressive decoding paradigm to present our method. When the block length equals the generation length, it becomes equivalent to full diffusion decoding.

\subsubsection{Drafting}
We denote the entire sequence as \( Y = [Y_{\text{left}}, {Y}_{\text{cur}}, Y_{\text{right}}] \), where \( Y_{\text{left}} \) contains the prompt \( X \) and the previously decoded blocks, \( {Y}_{\text{cur}} = [{y}_{\text{cur},1}, \ldots, {y}_{\text{cur},L_b}] \) represents the current block being decoded, and \( Y_\text{right} \) denotes the remaining blocks to be decoded. Here, \( L_b \) is the block length. At the $k$-th decoding step, instead of decoding a fixed number of tokens, we perform aggressive multi-token parallel decoding based on a confidence threshold $\tau_1$:
\begin{equation}
    {y}_{\text{cur},l}^{(k)} = \argmax\limits_{v \in V} p_\theta(\hat{{y}}_{\text{cur},l} = v|Y) , \text{ if } \max\limits_{v \in V} p_\theta(\hat{{y}}_{\text{cur},l} = v|Y) > \tau_1 \text{ and } {y}^{(k-1)}_{\text{cur},l}=\specialtoken{[MASK]}{gray}. 
\end{equation}

Here, a relatively low confidence threshold $\tau_1$ is adopted to allow more possible tokens to be decoded at each step, which will achieve the acceleration if only a few tokens among them are revoked during the verification module detailed in the next section. This will be demonstrated in the experiments.

\subsubsection{Verification}
The design principle of the verification module is to utilize the increasingly enriched semantic context at each decoding step—relative to earlier steps, to evaluate the quality of previously unmasked tokens. By re-masking low-quality tokens, the decoding process becomes revokable and amenable for the proper early error correction.

To realize effective quality verification about the decoded tokens, we design an auxiliary shadow block consisting entirely of \specialtoken{[MASK]}{gray}, $Y_{\text{shad}} = [\ \specialtoken{[MASK]}{gray}\ ]\times L_b$. This block is appended to the sequence $Y$, resulting in an extended sequence $\tilde{Y} = [Y_{\text{left}}, {Y}_{\text{cur}}, Y_{\text{right}},Y_{\text{shad}}]$. We carefully design the position IDs and attention mask associated with $Y_{\text{shad}}$ to ensure that its output can effectively verify the quality of the tokens decoded at the corresponding positions in $Y_{\text{cur}}$.

\textbf{Position IDs.} Although $Y_{\text{shad}}$ is appended to the right end of the sequence, we assign it the same position IDs as $Y_{\text{cur}}$. Thus, the output of $Y_{\text{shad}}$ corresponds to the same positions as $Y_{\text{cur}}$, enabling position-wise verification. 

\textbf{Attention Mask.} As illustrated in \cref{fig:attn_mask}, we carefully design the attention mask after incorporating $Y_{\text{shad}}$ into the sequence $\tilde{Y}$. Specifically, tokens in $Y_{\text{left}}$, $Y_{\text{cur}}$, and $Y_{\text{right}}$ can freely attend to each other, but they are not allowed to attend to $Y_{\text{shad}}$. In contrast, each token in $Y_{\text{shad}}$ is allowed to attend to all tokens except its corresponding position in $Y_{\text{cur}}$. 

With the above design of position IDs and attention masks, we achieve the following properties:
\begin{itemize}
    \item For any token in the current block $Y_{\text{cur}}$, appending $Y_{\text{shad}}$ does not affect the model's output. Formally,
    $$p_\theta(\hat{{y}}_{\text{cur},l} |Y) = p_\theta(\hat{{y}}_{\text{cur},l} |\tilde{Y}).
    $$
    
    \item For any token in $Y_{\text{shad}}$, the following properties hold. For example, consider the token $y_{\text{shad},3}$ in \cref{fig:method}, which is assigned position ID 6. 
    \begin{itemize}
	\item It shares the same position ID as $y_{\text{cur},3}$, and is allowed to attend to $Y_{\text{left}}$ and $Y_{\text{right}}$;
	\item It is explicitly prevented from attending to $y_{\text{cur},3}$, effectively avoiding information leakage during verification;
	\item For all other positions in $Y_{\text{cur}}$, each position is attended by exactly one decoded token (from $Y_{\text{cur}}$) and one \specialtoken{[MASK]}{gray} in $Y_{\text{shad}}$. The former provides progressively richer contextual semantics during decoding, while the latter serves to regularize the confidence of decoded tokens in $Y_{\text{cur}}$, reflecting the uncertainty and the need for potential refinement.
    \end{itemize}
\end{itemize}

With the specially designed position IDs and the attention mask described above, the verification module can be formally expressed as:
\begin{equation}
    {y}_{\text{cur},l}^{(k)} = \specialtoken{[MASK]}{gray}, \text{ if } p_\theta(\hat{{y}}_{\text{shad},l} = {y}_{\text{cur},l}^{(k-1)}|\tilde{Y})) < \tau_2 \text{ and } {y}_{\text{cur},l}^{(k-1)} \neq \specialtoken{[MASK]}{gray},
\end{equation}
where $\tau_2$ is the confidence threshold for verification.

\subsubsection{Overall Procedure}
In summary, at decoding step $k$, our framework enables both the drafting and verification processes to be completed in a single forward pass:
\begin{equation}
\label{eq: overall}
    {y}_{\text{cur},l}^{(k)} = 
\begin{cases}
\argmax\limits_{v \in V} p_\theta(\hat{{y}}_{\text{cur},l} = v|\tilde{Y}) , &\text{if } \max\limits_{v \in V} p_\theta(\hat{{y}}_{\text{cur},l} = v|\tilde{Y}) > \tau_1 \text{ and } {y}^{(k-1)}_{\text{cur},l}=\specialtoken{[MASK]}{gray}, \\
\specialtoken{[MASK]}{gray}, &\text{if } p_\theta(\hat{{y}}_{\text{shad},l} = {y}_{\text{cur},l}^{(k-1)}|\tilde{Y})) < \tau_2 \text{ and } {y}_{\text{cur},l}^{(k-1)} \neq \specialtoken{[MASK]}{gray}, \\
{y}_{\text{cur},l}^{(k-1)}, &\text{otherwise}.
\end{cases}
\end{equation}

 We iteratively refine the entire $Y_{\text{cur}}$ using the procedure in \cref{eq: overall}, until all tokens in $Y_{\text{cur}}$ are no longer \specialtoken{[MASK]}{gray}. We set the drafting threshold $\tau_1$ and the verification threshold $\tau_2$ such that $\tau_1 < \tau_2$. A lower $\tau_1$ accelerates the decoding process by allowing more tokens to be generated in parallel, while a higher $\tau_2$ ensures the quality of the final output by enforcing stricter acceptance criteria. We refer to this design philosophy as \emph{"Wide-In, Narrow-Out"} and term our method as {\method} in short.

\definecolor{light-gray}{gray}{0.6}
\colorlet{ourred}{red!4}
\colorlet{ourblue}{blue!100}
\begin{table}[t]
\centering
\caption{Performance and inference speedup comparison on diverse language benchmarks.}\vspace{-10pt}
\label{tab:llada-bench}
\resizebox{0.98\textwidth}{!}{
\begin{tabular}{@{}!{\vrule width 0pt}p{2.4cm}<{\centering}p{1.2cm}<{\centering}p{2.5cm}<{\centering}p{2.5cm}<{\centering}p{1.8cm}<{\centering}p{2.5cm}<{\centering}p{1.8cm}<{\centering}}
\toprule
\multirow{2}{*}{\textbf{Benchmark}} & \multirow{2}{*}{\textbf{Method}} & \multirow{2}{*}{\textbf{Accuracy}} & \multirow{2}{*}{\textbf{Steps}}  & {\textbf{Step}} & \multirow{2}{*}{\textbf{TPS}} & {\textbf{TPS}} \\
&&&& \textbf{Reduction} && \textbf{Speedup}\\
\midrule

\multirow{2}{*}{\shortstack{GSM8K \\ \scriptsize \color{light-gray}{Math Reasoning}}}        & LLaDA & 73.24 & 256 & 1.00 $\times$ & 17.76 & 1.00 $\times$ \\
                              & \cellcolor{ourred} {\method}  & \cellcolor{ourred} 75.82 (\textcolor{ourblue}{+2.58}) & \cellcolor{ourred} 41.93 (\textcolor{ourblue}{-214.07}) & 
                                 \cellcolor{ourred} 6.10 $\times$ & 
                                 \cellcolor{ourred} 100.53 (\textcolor{ourblue}{+82.77}) & 
                                 \cellcolor{ourred} 5.66 $\times$ \\
\midrule
\multirow{2}{*}{\shortstack{MATH-500 \\ \scriptsize \color{light-gray}{Math Reasoning}}}         & LLaDA & 32.00 & 256 & 1.00 $\times$ & 17.62 & 1.00 $\times$ \\
                              & \cellcolor{ourred} {\method}  & \cellcolor{ourred} 34.20 (\textcolor{ourblue}{+2.20}) & \cellcolor{ourred} 74.44 (\textcolor{ourblue}{-181.56}) & 
                                 \cellcolor{ourred} 3.44 $\times$ & 
                                 \cellcolor{ourred} 55.86 (\textcolor{ourblue}{+38.24}) & 
                                 \cellcolor{ourred} 3.17 $\times$ \\
\midrule
\multirow{2}{*}{\shortstack{HumanEval \\ \scriptsize \color{light-gray}{Code Generation}}}   & LLaDA & 37.80 & 256 & 1.00 $\times$ & 14.52 & 1.00 $\times$ \\
                              & \cellcolor{ourred} {\method}  & \cellcolor{ourred} 42.07 (\textcolor{ourblue}{+4.27}) & \cellcolor{ourred} 93.32 (\textcolor{ourblue}{-162.68}) & 
                                 \cellcolor{ourred} 2.74 $\times$ & 
                                 \cellcolor{ourred} 37.19 (\textcolor{ourblue}{+22.67}) & 
                                 \cellcolor{ourred} 2.56 $\times$ \\
\midrule
\multirow{2}{*}{\shortstack{MBPP \\ \scriptsize \color{light-gray}{Code Generation}}}        & LLaDA & 36.40 & 256 & 1.00 $\times$ & 18.52 & 1.00 $\times$ \\
                              & \cellcolor{ourred} {\method}  & \cellcolor{ourred} 36.40 (\textcolor{ourblue}{+0.00}) & \cellcolor{ourred} 96.57 (\textcolor{ourblue}{-159.43}) & 
                                 \cellcolor{ourred} 2.65 $\times$ & 
                                 \cellcolor{ourred} 45.39 (\textcolor{ourblue}{+26.87}) & 
                                 \cellcolor{ourred} 2.45 $\times$ \\
\midrule
\multirow{2}{*}{\shortstack{Countdown \\ \scriptsize \color{light-gray}{Logical Reasoning}}} & LLaDA & 24.21 & 256 & 1.00 $\times$ & 17.22 & 1.00 $\times$ \\
                              & \cellcolor{ourred} {\method}  & \cellcolor{ourred} 33.20 (\textcolor{ourblue}{+8.99}) & \cellcolor{ourred} 105.88 (\textcolor{ourblue}{-150.12}) & 
                                 \cellcolor{ourred} 2.41 $\times$ & 
                                 \cellcolor{ourred} 38.97 (\textcolor{ourblue}{+21.75}) & 
                                 \cellcolor{ourred} 2.26 $\times$ \\
\midrule
\multirow{2}{*}{\shortstack{Sudoku \\ \scriptsize \color{light-gray}{Logical Reasoning}}}    & LLaDA & 14.23 & 256 & 1.00 $\times$ & 11.61 & 1.00 $\times$ \\
                              & \cellcolor{ourred} {\method}  & \cellcolor{ourred} 15.20 (\textcolor{ourblue}{+0.97}) & \cellcolor{ourred} 131.96 (\textcolor{ourblue}{-124.04}) & 
                                 \cellcolor{ourred} 1.94 $\times$ & 
                                 \cellcolor{ourred} 21.11 (\textcolor{ourblue}{+9.50}) & 
                                 \cellcolor{ourred} 1.82 $\times$ \\
\midrule
\multirow{2}{*}{\shortstack{ARC-E \\ \scriptsize \color{light-gray}{Commonsense Reasoning}}}            & LLaDA & 59.13 & 256 & 1.00 $\times$ & 17.26 & 1.00 $\times$ \\
                              & \cellcolor{ourred} {\method}  & \cellcolor{ourred} 81.19 (\textcolor{ourblue}{+22.06}) & \cellcolor{ourred} 40.19 (\textcolor{ourblue}{-215.81}) & 
                                 \cellcolor{ourred} 6.37 $\times$ & 
                                 \cellcolor{ourred} 101.61 (\textcolor{ourblue}{+84.35}) & 
                                 \cellcolor{ourred} 5.89 $\times$ \\
\midrule
\multirow{2}{*}{\shortstack{ARC-C \\ \scriptsize \color{light-gray}{Commonsense Reasoning}}}            & LLaDA & 51.87 & 256 & 1.00 $\times$ & 17.10 & 1.00 $\times$ \\
                              & \cellcolor{ourred} {\method}  & \cellcolor{ourred} 73.89 (\textcolor{ourblue}{+22.02}) & \cellcolor{ourred} 47.41 (\textcolor{ourblue}{-208.59}) & 
                                 \cellcolor{ourred} 5.40 $\times$ & 
                                 \cellcolor{ourred} 85.42 (\textcolor{ourblue}{+68.32}) & 
                                 \cellcolor{ourred} 5.00 $\times$ \\
\bottomrule
\end{tabular}}
\vspace{-10pt}
\end{table}

\section{Experiment}
\subsection{Experiment setup}
\textbf{Datasets and Baselines.} We conduct experiment to evaluate {\method} across different types of tasks and domains. Specifically, for language domain, we compare {\method} with the standard decoding of LLaDA on eight tasks: GSM8K~\citep{DBLP:journals/corr/abs-2110-14168}, MATH-500~\citep{DBLP:conf/nips/HendrycksBKABTS21}, HumanEval~\citep{DBLP:journals/corr/abs-2107-03374}, MBPP~\citep{DBLP:journals/corr/abs-2108-07732}, Countdown~\citep{DBLP:journals/corr/abs-2504-12216}, Sudoku~\citep{DBLP:journals/corr/abs-2504-12216}, ARC-E~\citep{DBLP:journals/corr/abs-1803-05457}, and ARC-C~\citep{DBLP:journals/corr/abs-1803-05457}, covering four categories of textual generation tasks, including math reasoning, code generation, logical reasoning, and commonsense reasoning. For vision-language domain, we evaluate {\method} against the standard decoding of MMaDA~\citep{DBLP:journals/corr/abs-2505-15809} on six multimodal understanding tasks: Flickr30k~\citep{DBLP:journals/tacl/YoungLHH14}, AI2D~\citep{DBLP:journals/corr/KembhaviSKSHF16}, MATH-Vision~\citep{DBLP:conf/nips/WangPSLRZZL24}, MathVista~\citep{DBLP:conf/iclr/LuBX0LH0CG024}, MMMU~\citep{DBLP:conf/cvpr/YueNZ0LZSJRSWYY24} and ScienceQA~\citep{DBLP:conf/nips/LuMX0CZTCK22}, spanning four types of multimodal tasks---captioning, chart understanding, math reasoning and multi-discipline reasoning. For clarity, we test on the validation set of MMMU and the official testmini subset of MathVista. 

\textbf{Evaluation Details.} All benchmarks are evaluated in a zero-shot manner, except Sudoku, which is evaluated in a 4-shot setting. We use the CIDEr metric~\citep{DBLP:conf/cvpr/VedantamZP15} for the Flickr30k benchmark and accuracy for all the remaining benchmarks. To assess the inference efficiency of the decoding method, we measure the required decoding steps and Tokens Per Second (TPS) of the baselines and {\method} on every task by averaging over all the samples in a benchmark. 

\textbf{Implementation details.} We adopt the open-sourced LLaDA-8B-Instruct\footnote{\href{https://huggingface.co/GSAI-ML/LLaDA-8B-Instruct}{https://huggingface.co/GSAI-ML/LLaDA-8B-Instruct}} for language benchmarks and 
MMaDA-8B-MixCoT\footnote{\href{https://huggingface.co/Gen-Verse/MMaDA-8B-MixCoT}{https://huggingface.co/Gen-Verse/MMaDA-8B-MixCoT}} for vision-language tasks. 
We employ the semi-autoregressive sampling strategy introduced in LLaDA~\citep{DBLP:journals/corr/abs-2502-09992}, where the output sequence is partitioned into multiple blocks and generated from left to right. In our evaluation, we set the generation length to 256 and the block length to 128, unless specified otherwise. 
For the hyperparameters of {\method}, we set the verification threshold $\tau_{2}$ to 0.9 and tune the drafting threshold $\tau_{1}$ from \{0.5, 0.6, 0.7\}.

\subsection{Main Results}
\textbf{Performance and speedup on text generation.}
We report the performance, decoding steps and throughput (TPS) of LLaDA, with and without {\method}, on language benchmarks in \cref{tab:llada-bench}. {\method} achieves significantly better accuracy with far fewer decoding steps than the baseline LLaDA, except for the MBPP task, where {\method} achieves the same performance as LLaDA. For instance, {\method} improves accuracy on GSM8K by 2.58\% with 6.10$\times$ step reduction and 5.66$\times$ TPS speedup. 
\cref{tab:llada-bench} demonstrate the effectiveness of {\method} in enhancing generation quality and inference efficiency.

\begin{table}[t]
\centering
\caption{Performance and inference speedup comparison across diverse multi-modal understanding and reasoning benchmarks. We use CIDEr for Flickr30k and accuracy for other benchmarks.}\vspace{-10pt}
\label{tab:mmada-bench}
\resizebox{0.98\textwidth}{!}{
\begin{tabular}{@{}!{\vrule width 0pt}p{2.5cm}<{\centering}p{1.2cm}<{\centering}p{2.5cm}<{\centering}p{2.5cm}<{\centering}p{1.8cm}<{\centering}p{2.5cm}<{\centering}p{1.8cm}<{\centering}}
\toprule
\multirow{2}{*}{\textbf{Benchmark}} & \multirow{2}{*}{\textbf{Method}} & \multirow{2}{*}{\textbf{Performance}} & \multirow{2}{*}{\textbf{Steps}}  & {\textbf{Step}} & \multirow{2}{*}{\textbf{TPS}} & {\textbf{TPS}} \\
&&&& \textbf{Reduction} && \textbf{Speedup}\\
\midrule 
\multirow{2}{*}{\shortstack{Flickr30k \\ \scriptsize \color{light-gray}{Captioning}}}                  & MMaDA & 53.67 & 256 & 1.00 $\times$ & 6.41 & 1.00 $\times$ \\
       & \cellcolor{ourred} {\method}  & \cellcolor{ourred} 53.83 (\textcolor{ourblue}{+0.16}) & \cellcolor{ourred} 25.47 (\textcolor{ourblue}{-230.53}) & 
                                 \cellcolor{ourred} 10.05 $\times$ & 
                                 \cellcolor{ourred} 55.11 (\textcolor{ourblue}{+48.70}) & 
                                 \cellcolor{ourred} 8.60 $\times$ \\
\midrule
\multirow{2}{*}{\shortstack{AI2D \\ \scriptsize \color{light-gray}{Chart Understanding}}}              & MMaDA & 54.86 & 256 & 1.00 $\times$ & 6.31 & 1.00 $\times$  \\
        & \cellcolor{ourred} {\method}  & \cellcolor{ourred} 57.19 (\textcolor{ourblue}{+2.33}) & \cellcolor{ourred} 30.90 (\textcolor{ourblue}{-225.10}) & 
                                 \cellcolor{ourred} 8.30 $\times$ & 
                                 \cellcolor{ourred} 46.04 (\textcolor{ourblue}{+39.73}) & 
                                 \cellcolor{ourred} 7.30 $\times$ \\
\midrule
\multirow{2}{*}{\shortstack{MATH-Vision \\ \scriptsize \color{light-gray}{Math Reasoning}}}            & MMaDA & 8.55  & 256 & 1.00 $\times$ & 6.22 & 1.00 $\times$ \\
        & \cellcolor{ourred} {\method}  & \cellcolor{ourred} 9.57 (\textcolor{ourblue}{+1.02}) & \cellcolor{ourred} 44.69 (\textcolor{ourblue}{-211.31}) & 
                                 \cellcolor{ourred} 5.73 $\times$ & 
                                 \cellcolor{ourred} 31.17 (\textcolor{ourblue}{+24.95}) & 
                                 \cellcolor{ourred} 5.01 $\times$ \\
\midrule
\multirow{2}{*}{\shortstack{MathVista-mini \\ \scriptsize \color{light-gray}{Math Reasoning}}}         & MMaDA & 31.10 & 256 & 1.00 $\times$ & 6.21 & 1.00 $\times$ \\
        & \cellcolor{ourred} {\method}  & \cellcolor{ourred} 31.40 (\textcolor{ourblue}{+0.30}) & \cellcolor{ourred} 33.45 (\textcolor{ourblue}{-222.55}) & 
                                 \cellcolor{ourred} 7.65 $\times$ & 
                                 \cellcolor{ourred} 41.96 (\textcolor{ourblue}{+35.75}) & 
                                 \cellcolor{ourred} 6.76 $\times$ \\
\midrule
\multirow{2}{*}{\shortstack{MMMU-val \\ \scriptsize \color{light-gray}{Multi-discipline Reasoning}}}   & MMaDA & 18.56 & 256 & 1.00 $\times$ & 6.02 & 1.00 $\times$ \\
        & \cellcolor{ourred} {\method}  & \cellcolor{ourred} 24.00 (\textcolor{ourblue}{+5.44}) & \cellcolor{ourred} 38.47 (\textcolor{ourblue}{-217.53}) & 
                                 \cellcolor{ourred} 6.65 $\times$ & 
                                 \cellcolor{ourred} 36.13 (\textcolor{ourblue}{+30.11}) & 
                                 \cellcolor{ourred} 6.00 $\times$ \\
\midrule
\multirow{2}{*}{\shortstack{ScienceQA \\ \scriptsize \color{light-gray}{Multi-discipline Reasoning}}}  & MMaDA & 30.89 & 256 & 1.00 $\times$ & 6.07 & 1.00 $\times$ \\
        & \cellcolor{ourred} {\method}  & \cellcolor{ourred} 42.24 (\textcolor{ourblue}{+11.35}) & \cellcolor{ourred} 28.12 (\textcolor{ourblue}{-227.88}) & 
                                 \cellcolor{ourred} 9.10 $\times$ & 
                                 \cellcolor{ourred} 49.45 (\textcolor{ourblue}{+43.38}) & 
                                 \cellcolor{ourred} 8.15 $\times$ \\
\bottomrule
\end{tabular}}
\vspace{-5pt}
\end{table}

\textbf{Performance and speedup on multimodal understanding and reasoning.} We assess the performance and efficiency gain of {\method} incorporated into MMaDA and summarize the results in \cref{tab:mmada-bench}. Compared to the vanilla MMaDA, {\method} demonstrates consistent and substantial improvements in inference efficiency across all benchmarks. Notably, the speedup effect is even more pronounced than that on textual domain tasks, when compared with results in \cref{tab:llada-bench}. Furthermore, {\method} greatly improves the task performance of MMaDA on AI2D, MMMU and ScienceQA while maintaining comparable results on Flickr30k, MATH-Vision and MathVista. These results indicate that {\method} consistently delivers both performance gains and accelerated inference in the multimodal domain.

\begin{wrapfigure}{r}{0.38\textwidth}
    \vspace*{-15pt}
    \centering
    \includegraphics[width=0.96\linewidth]{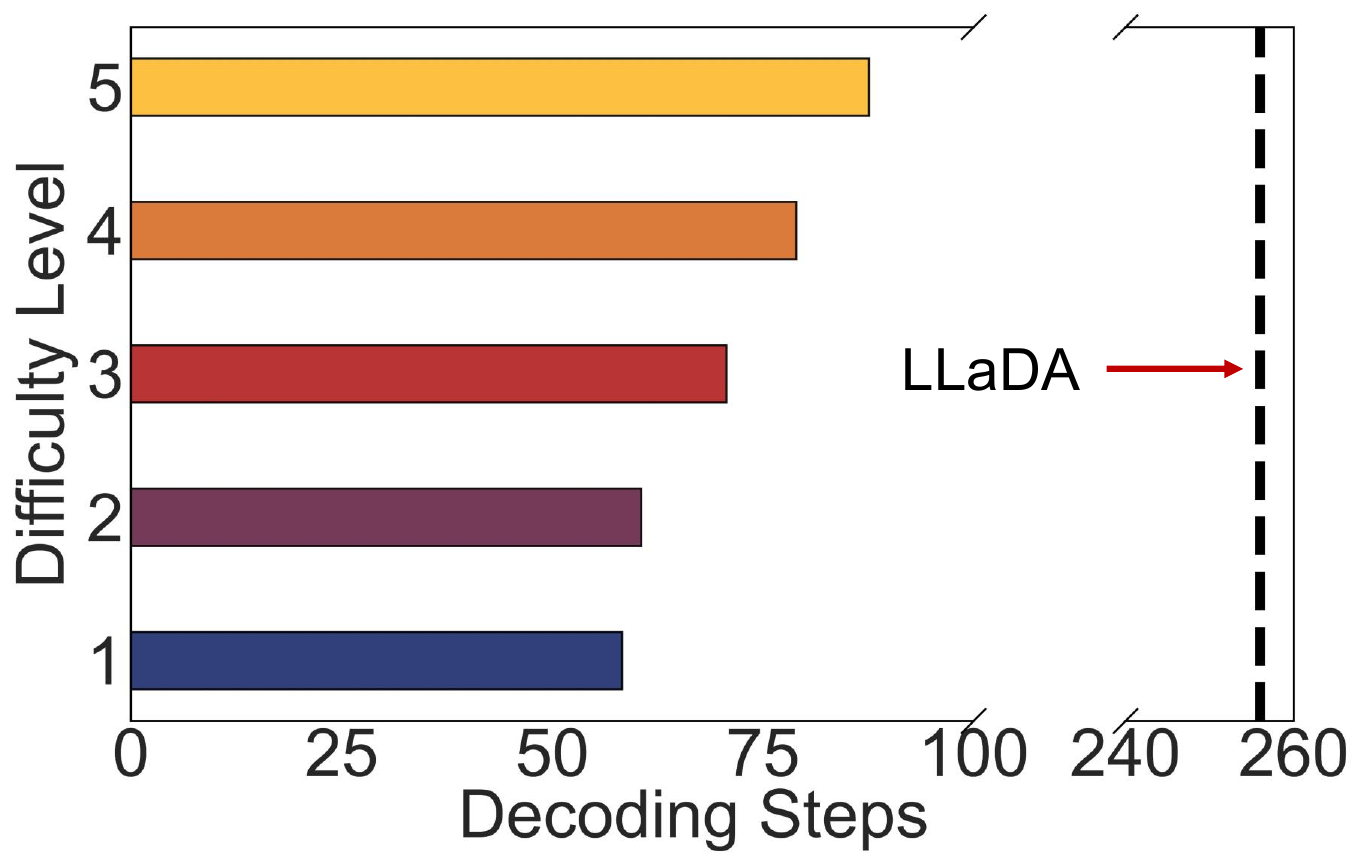}
    \vspace{-10pt}
    \caption{Decoding steps of {\method} on subsets of the MATH benchmark with varied difficulty levels.}
    \vspace{-10pt}
    \label{fig:difficulty_level}
\end{wrapfigure}

\textbf{Relation between speedup and task complexity.} As shown in \cref{tab:llada-bench} and \cref{tab:mmada-bench},  we observe a consistent positive correlation between the degree of speedup and task performance across all benchmarks. 
For instance, {\method} achieves a 10.05$\times$ step reduction on the relatively simple captioning task Flickr30k, compared to only 5.73$\times$ step reduction on the more challenging math reasoning benchmark MATH-Vision. 
This is because models can, in principle, solve tasks they are more proficient at with  lower computational cost, leaving greater room for acceleration under our decoding method.
And since models are typically more confident when handling easier tasks, each decoding step in {\method} tends to yield a larger number of effective tokens.
To further investigate this, we evaluate the decoding steps of {\method} across subsets of the MATH-500 benchmark categorized by difficulty levels. As shown in \cref{fig:difficulty_level}, {\method} achieves progressively greater acceleration as the difficulty decreases, highlighting its capability to adaptively optimize inference speed based on task complexity.

\subsection{Ablation Study and Further Analysis.}
\textbf{On different generation length.} In \cref{tab:gen_length&full_diff}, we evaluate the performance of {\method} with experiments on different generation lengths, where the block length $L_b$ is fixed to 128 and the baselines unmask 1 token every decoding step (to achieve their best generation performance). When the generation length is set to 512, {\method} still achieves comparable or better task performance with significantly fewer decoding steps, demonstrating the effectiveness of {\method} across different generation lengths.

\begin{table}[t]
\centering
\caption{Experiment results on different generation lengths and full diffusion setting, respectively. }\vspace{-10pt}
\label{tab:gen_length&full_diff}
\resizebox{0.98\textwidth}{!}{
\begin{tabular}{@{}!{\vrule width 0pt}p{1.5cm}<{\centering}p{1.5cm}<{\centering}p{1.5cm}<{\centering}p{1.2cm}<{\centering}p{2.5cm}<{\centering}p{1.5cm}<{\centering}p{1.5cm}<{\centering}p{1.5cm}<{\centering}p{1.5cm}<{\centering}}
\toprule
\multirow{2}{*}{\textbf{Benchmark}} & \textbf{Generation} & \textbf{Block} & \multirow{2}{*}{\textbf{Method}} & \multirow{2}{*}{\textbf{Accuracy}} & \multirow{2}{*}{\textbf{Steps}}  & {\textbf{Step}} & \multirow{2}{*}{\textbf{TPS}} & {\textbf{TPS}} \\
& \textbf{Length} & \textbf{Length} & & & & \textbf{Reduction} & & \textbf{Speedup}\\
\midrule
\multicolumn{9}{c}{\emph{Different Generation Lengths}} \\
\multirow{4}{*}{GSM8K}               & \multirow{2}{*}{256} & \multirow{2}{*}{128} & LLaDA     & 73.24 & 256   & 1.00 $\times$ & 17.76  & 1.00 $\times$        \\
                                     &                      & & \cellcolor{ourred} {\method} & \cellcolor{ourred} 75.82 (\textcolor{ourblue}{+2.58}) & \cellcolor{ourred} 41.93 &\cellcolor{ourred} 6.10 $\times$ &\cellcolor{ourred} 100.53 &\cellcolor{ourred} 5.66 $\times$      \\
                                     & \multirow{2}{*}{512} & \multirow{2}{*}{128} & LLaDA     & 74.60 & 512   & 1.00 $\times$ & 11.84 & 1.00 $\times$     \\
                                     &                      & & \cellcolor{ourred} {\method} &\cellcolor{ourred} 79.91 (\textcolor{ourblue}{+5.31}) &\cellcolor{ourred} 68.53 &\cellcolor{ourred} 7.47 $\times$ &\cellcolor{ourred} 82.64 &\cellcolor{ourred} 6.98 $\times$     \\
\midrule
\multirow{4}{*}{MMMU-val}            & \multirow{2}{*}{256} & \multirow{2}{*}{128} & MMaDA     & 18.56 & 256   & 1.00 $\times$ & 6.02  & 1.00 $\times$ \\
                                     &                      & &\cellcolor{ourred} {\method} &\cellcolor{ourred} 24.00 (\textcolor{ourblue}{+5.44}) &\cellcolor{ourred} 38.47 &\cellcolor{ourred} 6.65 $\times$ &\cellcolor{ourred} 36.13 &\cellcolor{ourred} 6.00 $\times$ \\
                                     & \multirow{2}{*}{512} & \multirow{2}{*}{128} & MMaDA     & 18.44 & 512 & 1.00 $\times$ & 5.01 & 1.00 $\times$     \\
                                     &                      & &\cellcolor{ourred} {\method} &\cellcolor{ourred} 23.44 
                                     (\textcolor{ourblue}{+5.00}) &\cellcolor{ourred} 64.82 &\cellcolor{ourred} 7.90 $\times$ &\cellcolor{ourred} 35.01 &\cellcolor{ourred} 6.99 $\times$  \\
\midrule\midrule
\multicolumn{9}{c}{\emph{Full Diffusion}} \\
\multirow{4}{*}{GSM8K}               & \multirow{2}{*}{256}  & \multirow{2}{*}{256} & LLaDA     & 34.34 & 256   & 1.00 $\times$ & 17.73 & 1.00 $\times$      \\
                                     &                       & &\cellcolor{ourred} {\method}   &\cellcolor{ourred} 58.22 (\textcolor{ourblue}{+23.88})  &\cellcolor{ourred} 38.77 &\cellcolor{ourred} 6.60 $\times$ &\cellcolor{ourred} 93.61 &\cellcolor{ourred} 5.28 $\times$  \\
                                     & \multirow{2}{*}{128}  & \multirow{2}{*}{128} & LLaDA     & 58.60 & 128   & 1.00 $\times$ & 23.23 & 1.00 $\times$  \\
                                     &                       & &\cellcolor{ourred} {\method}   &\cellcolor{ourred} 62.32 (\textcolor{ourblue}{+3.72}) &\cellcolor{ourred} 23.95 &\cellcolor{ourred} 5.34 $\times$ &\cellcolor{ourred} 114.29 &\cellcolor{ourred} 4.92 $\times$  \\
\midrule
\multirow{4}{*}{MMMU-val}            & \multirow{2}{*}{256}  & \multirow{2}{*}{256} & MMaDA     & 17.22 & 256   & 1.00$\times$ & 6.11 & 1.00$\times$   \\
                                     &                       & &\cellcolor{ourred} {\method} &\cellcolor{ourred} 22.44 (\textcolor{ourblue}{+5.22}) &\cellcolor{ourred} 24.94 &\cellcolor{ourred} 10.26$\times$ &\cellcolor{ourred} 50.03 &\cellcolor{ourred} 8.19$\times$    \\
                                     & \multirow{2}{*}{128}  & \multirow{2}{*}{128} & MMaDA     & 15.33 & 128   & 1.00$\times$  & 6.70 & 1.00 $\times$    \\
                                     &                       & & \cellcolor{ourred} \method &\cellcolor{ourred} 23.11 (\textcolor{ourblue}{+7.78}) &\cellcolor{ourred} 19.14 &\cellcolor{ourred} 6.69$\times$  &\cellcolor{ourred} 39.94 &\cellcolor{ourred} 5.96 $\times$    \\
\bottomrule
\end{tabular}
}\vspace{-5pt}
\end{table}
\begin{table}[t]
\centering
\caption{Experiment results on the variant of {\method} without the verification module.}\vspace{-10pt}
\label{tab:verification_module}
\resizebox{0.98\textwidth}{!}{
\begin{tabular}{@{}!{\vrule width 0pt}p{2.5cm}<{\centering}p{3.5cm}<{\centering}p{1.5cm}<{\centering}p{1.5cm}<{\centering}p{1.5cm}<{\centering}p{1.5cm}<{\centering}p{1.5cm}<{\centering}}
\toprule
\multirow{2}{*}{\textbf{Benchmark}} & \multirow{2}{*}{\textbf{Method}} & \multirow{2}{*}{\textbf{Accuracy}} & \multirow{2}{*}{\textbf{Steps}}  & {\textbf{Step}} & \multirow{2}{*}{\textbf{TPS}} & {\textbf{TPS}} \\
& & & & \textbf{Reduction} & & \textbf{Speedup}\\
\midrule
\multirow{4}{*}{GSM8K}      & LLaDA   & 73.24 & 256 & 1.00 $\times$ & 17.76 & 1.00 $\times$ \\     
                            & Only Draft ($\tau_1=0.6$) & 70.28 & 34.79 & 7.36 $\times$ & 130.89 & 7.37 $\times$       \\
                            & Only Draft ($\tau_1=0.9$) & 72.33 & 81.39 & 3.15 $\times$ & 56.12 & 3.16 $\times$       \\
                            & \method                   & 75.82 & 41.93 & 6.10 $\times$ & 100.53 & 5.66 $\times$      \\
\midrule
\multirow{4}{*}{MMMU-val}   & MMaDA & 18.56 & 256 & 1.00 $\times$ & 6.02 & 1.00 $\times$ \\
                            & Only Draft ($\tau_1=0.6$) & 19.89 & 35.63 & 7.18 $\times$ & 43.22 & 7.18 $\times$      \\
                            & Only Draft ($\tau_1=0.9$) & 18.56 & 79.74 & 3.21 $\times$ & 19.38 & 3.22 $\times$       \\
                            & \method                   & 24.00 & 38.47 & 6.65 $\times$ & 36.13 & 6.00 $\times$ \\
\bottomrule
\end{tabular}
}\vspace{-5pt}
\end{table}

\textbf{On full diffusion decoding (instead of semi-autoregressive decoding).} In \cref{tab:gen_length&full_diff}, we compare the baselines and {\method} applying full diffusion decoding, which means the block length $L_b$ is set equal to the generation length. Compared to results on the semi-autoregressive decoding in \cref{tab:llada-bench} and \cref{tab:mmada-bench}, {\method} demonstrates substantially strong accuracy gains under the full diffusion setting. Notably, while LLaDA suffers a substantial accuracy drop on GSM8K with full diffusion decoding, {\method} maintains reasonable performance with far fewer decoding steps. These results indicate that {\method} unlocks significantly greater potential for boosting model performance and computational efficiency when applied in full diffusion decoding scenarios.

\textbf{Comparison with naive parallel sampling.} The decoding process of existing DLLMs can be sped up by generating multiple tokens per step, \emph{i.e.}, naive parallel sampling. However, directly increasing the fixed number of generated tokens per step for DLLMs leads to significant performance degradation. For instance, on GSM8K, accuracy drops from 73.24\% with 256 steps (1 token/step) to 71.11\% with 128 steps (2 tokens/step), and further down to 64.67\% with 64 steps (4 tokens/step). In contrast, the draft-and-verify procedure of {\method} enables flexible decoding during the generation process, achieving 75.82\% accuracy with only 41.93 steps on average, corresponding to a 6.10$\times$ speedup, thereby substantially improving task performance while accelerating inference.

\textbf{Ablation on verification module.} We conduct an ablation study on a variant of {\method} that excludes the verification module, implemented by setting the verification threshold $\tau_2$ to zero. As presented in \cref{tab:verification_module}, this variant exhibits significant performance degradation across both benchmarks compared to {\method}. Specifically, when the drafting threshold $\tau_1$ is small (corresponding to 0.6 in the table), more candidate tokens are unmasked per decoding step, which naturally introduces a higher proportion of unreliable tokens and ultimately compromises output quality. Conversely, when $\tau_1$ is large (corresponding to 0.9 in the table), fewer candidate tokens are unmasked per decoding step, which can mitigate error propagation but at the expense of computational efficiency. Crucially, without the verification module, the generation process lacks a mechanism to correct erroneous predictions. As a result, even with a large $\tau_1$, the model may fail to achieve comparable performance, underscoring the necessity of the verification module in maintaining both generation quality. 

\begin{figure*}[!t]
\centering 
\vspace{-5pt}
\subfigure[\small {\method} (LLaDA-based) on GSM8K.]{
\begin{minipage}{0.48\textwidth}
\centering
\includegraphics[width=\textwidth]{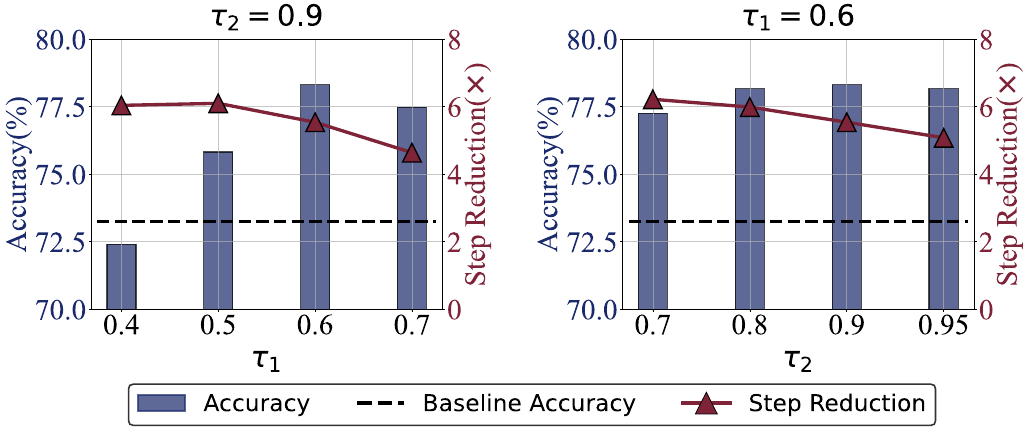}
\label{fig:ab_tau_language}
\end{minipage} 
}
\subfigure[\small {\method} (MMaDA-based) on MMMU-val.]{
\begin{minipage}{0.48\textwidth}
\centering
\includegraphics[width=\textwidth]{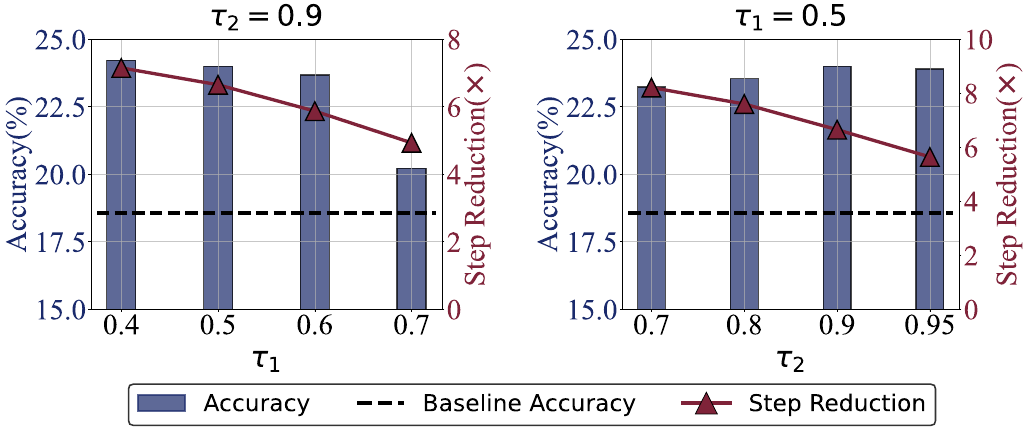}
\label{fig:ab_tau_mmmu}
\end{minipage} 
}
\vspace{-10pt}
\caption{Ablation study on the drafting threshold $\tau_1$ and the verification threshold $\tau_2$.}
\label{fig:ab_tuning_tau}
\vspace{-10pt}
\end{figure*}

\begin{figure}
    \centering
    \includegraphics[width=0.97\linewidth]{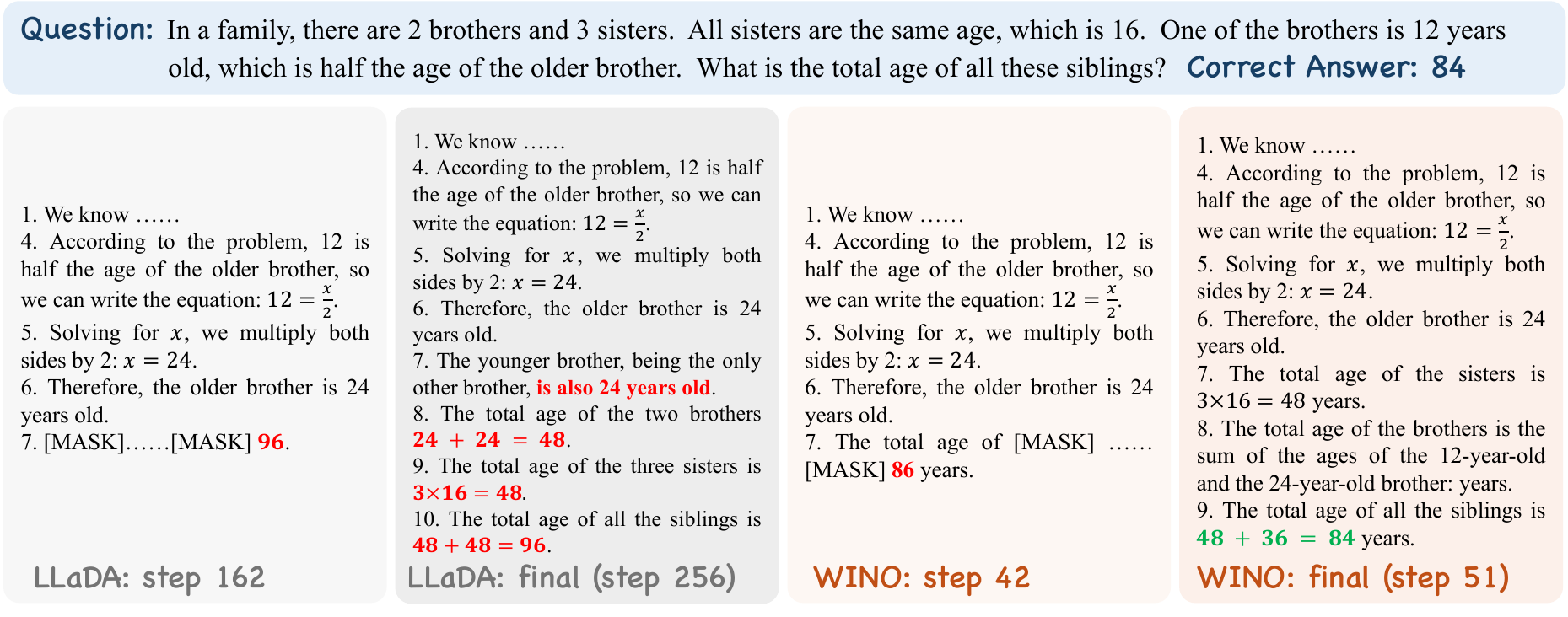}
    \vspace{-10pt}
    \caption{Case Study: GSM8K Example. We compare standard decoding with LLaDA against the intermediate and final results produced by WINO. More detailed case studies are provided in \cref{appendix: case}.}
    \vspace{-13pt}
    \label{fig: case-gsm8k}
\end{figure}

\textbf{Effect of threshold tuning.} In \cref{fig:ab_tuning_tau}, we present the evaluation results of {\method} with varying drafting threshold $\tau_1$ and  verification threshold $\tau_2$. Our experiments suggest that {\method} consistently outperforms baselines across different benchmarks and the $\tau_1$ and $\tau_2$ values in terms of both task performance and inference efficiency. As the $\tau_1$ value decreases, more candidate tokens are unmasked at each decoding step, thereby accelerating inference by reducing the required decoding steps. However, this comes at the cost of introducing more unreliable predictions, which may place a greater burden on the verification module to correct errors. Empirically, we find that setting the value of drafting threshold $\tau_1$ within the range of 0.5 to 0.7 achieves an optimal balance, maintaining competitive task performance while preserving efficient generation. The verification threshold $\tau_2$ controls the strictness of the verification process and thus influences decoding speed. Since the performance is relatively robust to $\tau_2$, we fix $\tau_2 = 0.9$ in all experiments, while leaving open the possibility of further tuning this parameter for even better performance and speedup.

\begin{wrapfigure}{r}{0.34\textwidth}
    \vspace*{-8pt}
    \centering
    \includegraphics[width=0.9\linewidth]{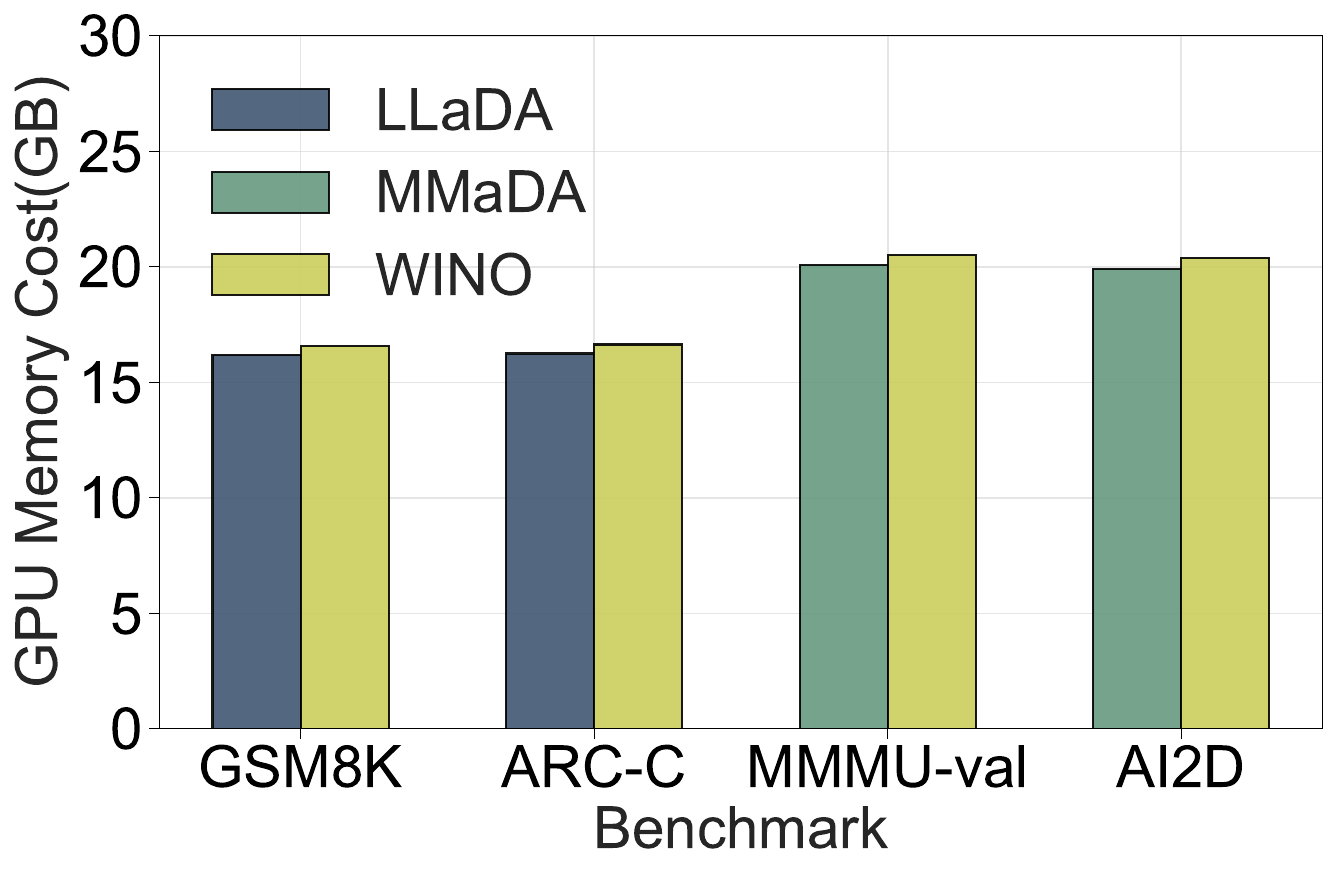}
    \vspace{-8pt}
    \caption{GPU memory usage.}
    \vspace{-10pt}
    \label{fig:gpu_mem}
\end{wrapfigure}
\textbf{GPU memory usage.} To facilitate efficient and effective quality verification of the unmasked tokens, {\method} introduces an auxiliary shadow block whose size equals the specified block length $L_b$ of the semi-autoregressive decoding process. Therefore, {\method} induces additional GPU memory cost due to the longer sequence length. We evaluate the GPU memory cost of {\method} and present the results in \cref{fig:gpu_mem}. The additional memory cost of {\method} remains marginal compared to the baselines across all the benchmarks. For instance, on GSM8K, {\method} increases GPU memory usage by only 2.4\% (from 16.18 to 16.57 GB) compared to standard LLaMA decoding. For other datasets, the memory overhead introduced by WINO is similarly negligible.

\textbf{Case Study: Decoding Dynamics.} To conduct a fine-grained examination of the decoding processes of {\method}, we present an example from GSM8K in \cref{fig: case-gsm8k}. As shown, the baseline may produce erroneous tokens at the early decoding stages. Since the generated tokens by the baseline remain unchanged in subsequent decoding steps, the false contextual information propagates throughout the whole generation process, eventually leading to low-quality generation results. In contrast, {\method} enables dynamical refinement of generated tokens via an iterative draft-and-verify mechanism, which mitigates error accumulation and facilitates high-quality decoded outputs.

\section{Conclusion}
In this work, we introduce {\Method} ({\method}), a training-free decoding algorithm that resolves the critical quality-speed trade-off in Diffusion Large Language Models (DLLMs) by making their generation process revokable. {\method} overcomes the limitations of irrevokable standard decoding by employing a parallel draft-and-verify mechanism, allowing the model to aggressively generate tokens and iteratively correct errors using its full bidirectional context. Our experiments on existing open-source models like LLaDA and MMaDA demonstrate that {\method} simultaneously accelerates inference by up to 10$\times$ while significantly improving accuracy across a diverse set of language and vision-language tasks. While acknowledging areas for future architectural improvements, {\method} fundamentally enhances the practicality of DLLMs by rethinking the decoding process itself, establishing them as a truly efficient and high-quality alternative to autoregressive systems.

{
\bibliography{main}

\begin{thebibliography}{41}
\providecommand{\natexlab}[1]{#1}
\providecommand{\url}[1]{\texttt{#1}}
\expandafter\ifx\csname urlstyle\endcsname\relax
  \providecommand{\doi}[1]{doi: #1}\else
  \providecommand{\doi}{doi: \begingroup \urlstyle{rm}\Url}\fi

\bibitem[Arriola et~al.(2025)Arriola, Gokaslan, Chiu, Yang, Qi, Han, Sahoo, and Kuleshov]{DBLP:conf/iclr/ArriolaGCYQHSK25}
Marianne Arriola, Aaron Gokaslan, Justin~T. Chiu, Zhihan Yang, Zhixuan Qi, Jiaqi Han, Subham~Sekhar Sahoo, and Volodymyr Kuleshov.
\newblock Block diffusion: Interpolating between autoregressive and diffusion language models.
\newblock In \emph{The Thirteenth International Conference on Learning Representations, {ICLR} 2025, Singapore, April 24-28, 2025}. OpenReview.net, 2025.

\bibitem[Austin et~al.(2021{\natexlab{a}})Austin, Johnson, Ho, Tarlow, and van~den Berg]{DBLP:conf/nips/AustinJHTB21}
Jacob Austin, Daniel~D. Johnson, Jonathan Ho, Daniel Tarlow, and Rianne van~den Berg.
\newblock Structured denoising diffusion models in discrete state-spaces.
\newblock In Marc'Aurelio Ranzato, Alina Beygelzimer, Yann~N. Dauphin, Percy Liang, and Jennifer~Wortman Vaughan, editors, \emph{Advances in Neural Information Processing Systems 34: Annual Conference on Neural Information Processing Systems 2021, NeurIPS 2021, December 6-14, 2021, virtual}, pages 17981--17993, 2021{\natexlab{a}}.

\bibitem[Austin et~al.(2021{\natexlab{b}})Austin, Odena, Nye, Bosma, Michalewski, Dohan, Jiang, Cai, Terry, Le, and Sutton]{DBLP:journals/corr/abs-2108-07732}
Jacob Austin, Augustus Odena, Maxwell~I. Nye, Maarten Bosma, Henryk Michalewski, David Dohan, Ellen Jiang, Carrie~J. Cai, Michael Terry, Quoc~V. Le, and Charles Sutton.
\newblock Program synthesis with large language models.
\newblock \emph{CoRR}, abs/2108.07732, 2021{\natexlab{b}}.

\bibitem[Ben{-}Hamu et~al.(2025)Ben{-}Hamu, Gat, Severo, Nolte, and Karrer]{DBLP:journals/corr/abs-2505-24857}
Heli Ben{-}Hamu, Itai Gat, Daniel Severo, Niklas Nolte, and Brian Karrer.
\newblock Accelerated sampling from masked diffusion models via entropy bounded unmasking.
\newblock \emph{CoRR}, abs/2505.24857, 2025.

\bibitem[Campbell et~al.(2022)Campbell, Benton, Bortoli, Rainforth, Deligiannidis, and Doucet]{DBLP:conf/nips/CampbellBBRDD22}
Andrew Campbell, Joe Benton, Valentin~De Bortoli, Thomas Rainforth, George Deligiannidis, and Arnaud Doucet.
\newblock A continuous time framework for discrete denoising models.
\newblock In Sanmi Koyejo, S.~Mohamed, A.~Agarwal, Danielle Belgrave, K.~Cho, and A.~Oh, editors, \emph{Advances in Neural Information Processing Systems 35: Annual Conference on Neural Information Processing Systems 2022, NeurIPS 2022, New Orleans, LA, USA, November 28 - December 9, 2022}, 2022.

\bibitem[Chen et~al.(2021)Chen, Tworek, Jun, Yuan, de~Oliveira~Pinto, Kaplan, Edwards, Burda, Joseph, Brockman, Ray, Puri, Krueger, Petrov, Khlaaf, Sastry, Mishkin, Chan, Gray, Ryder, Pavlov, Power, Kaiser, Bavarian, Winter, Tillet, Such, Cummings, Plappert, Chantzis, Barnes, Herbert{-}Voss, Guss, Nichol, Paino, Tezak, Tang, Babuschkin, Balaji, Jain, Saunders, Hesse, Carr, Leike, Achiam, Misra, Morikawa, Radford, Knight, Brundage, Murati, Mayer, Welinder, McGrew, Amodei, McCandlish, Sutskever, and Zaremba]{DBLP:journals/corr/abs-2107-03374}
Mark Chen, Jerry Tworek, Heewoo Jun, Qiming Yuan, Henrique~Pond{\'{e}} de~Oliveira~Pinto, Jared Kaplan, Harri Edwards, Yuri Burda, Nicholas Joseph, Greg Brockman, Alex Ray, Raul Puri, Gretchen Krueger, Michael Petrov, Heidy Khlaaf, Girish Sastry, Pamela Mishkin, Brooke Chan, Scott Gray, Nick Ryder, Mikhail Pavlov, Alethea Power, Lukasz Kaiser, Mohammad Bavarian, Clemens Winter, Philippe Tillet, Felipe~Petroski Such, Dave Cummings, Matthias Plappert, Fotios Chantzis, Elizabeth Barnes, Ariel Herbert{-}Voss, William~Hebgen Guss, Alex Nichol, Alex Paino, Nikolas Tezak, Jie Tang, Igor Babuschkin, Suchir Balaji, Shantanu Jain, William Saunders, Christopher Hesse, Andrew~N. Carr, Jan Leike, Joshua Achiam, Vedant Misra, Evan Morikawa, Alec Radford, Matthew Knight, Miles Brundage, Mira Murati, Katie Mayer, Peter Welinder, Bob McGrew, Dario Amodei, Sam McCandlish, Ilya Sutskever, and Wojciech Zaremba.
\newblock Evaluating large language models trained on code.
\newblock \emph{CoRR}, abs/2107.03374, 2021.

\bibitem[Clark et~al.(2018)Clark, Cowhey, Etzioni, Khot, Sabharwal, Schoenick, and Tafjord]{DBLP:journals/corr/abs-1803-05457}
Peter Clark, Isaac Cowhey, Oren Etzioni, Tushar Khot, Ashish Sabharwal, Carissa Schoenick, and Oyvind Tafjord.
\newblock Think you have solved question answering? try arc, the {AI2} reasoning challenge.
\newblock \emph{CoRR}, abs/1803.05457, 2018.

\bibitem[Cobbe et~al.(2021)Cobbe, Kosaraju, Bavarian, Chen, Jun, Kaiser, Plappert, Tworek, Hilton, Nakano, Hesse, and Schulman]{DBLP:journals/corr/abs-2110-14168}
Karl Cobbe, Vineet Kosaraju, Mohammad Bavarian, Mark Chen, Heewoo Jun, Lukasz Kaiser, Matthias Plappert, Jerry Tworek, Jacob Hilton, Reiichiro Nakano, Christopher Hesse, and John Schulman.
\newblock Training verifiers to solve math word problems.
\newblock \emph{CoRR}, abs/2110.14168, 2021.

\bibitem[{Google DeepMind}(2025)]{deepmind2025geminidiffusion}
{Google DeepMind}.
\newblock Gemini diffusion.
\newblock Google DeepMind Models, 2025.
\newblock URL \url{https://deepmind.google/models/gemini-diffusion}.

\bibitem[Hendrycks et~al.(2021)Hendrycks, Burns, Kadavath, Arora, Basart, Tang, Song, and Steinhardt]{DBLP:conf/nips/HendrycksBKABTS21}
Dan Hendrycks, Collin Burns, Saurav Kadavath, Akul Arora, Steven Basart, Eric Tang, Dawn Song, and Jacob Steinhardt.
\newblock Measuring mathematical problem solving with the {MATH} dataset.
\newblock In \emph{NeurIPS Datasets and Benchmarks}, 2021.

\bibitem[Ho et~al.(2020)Ho, Jain, and Abbeel]{DBLP:conf/nips/HoJA20}
Jonathan Ho, Ajay Jain, and Pieter Abbeel.
\newblock Denoising diffusion probabilistic models.
\newblock In Hugo Larochelle, Marc'Aurelio Ranzato, Raia Hadsell, Maria{-}Florina Balcan, and Hsuan{-}Tien Lin, editors, \emph{Advances in Neural Information Processing Systems 33: Annual Conference on Neural Information Processing Systems 2020, NeurIPS 2020, December 6-12, 2020, virtual}, 2020.

\bibitem[{Inception Labs}(2025)]{inceptionlabs2025mercury}
{Inception Labs}.
\newblock Introducing mercury: The first commercial diffusion-based language model.
\newblock Inception Labs Blog, 2025.
\newblock URL \url{https://www.inceptionlabs.ai/introducing-mercury}.

\bibitem[Kembhavi et~al.(2016)Kembhavi, Salvato, Kolve, Seo, Hajishirzi, and Farhadi]{DBLP:journals/corr/KembhaviSKSHF16}
Aniruddha Kembhavi, Mike Salvato, Eric Kolve, Min~Joon Seo, Hannaneh Hajishirzi, and Ali Farhadi.
\newblock A diagram is worth {A} dozen images.
\newblock \emph{CoRR}, abs/1603.07396, 2016.

\bibitem[Li et~al.(2022)Li, Thickstun, Gulrajani, Liang, and Hashimoto]{DBLP:conf/nips/LiTGLH22}
Xiang~Lisa Li, John Thickstun, Ishaan Gulrajani, Percy Liang, and Tatsunori~B. Hashimoto.
\newblock Diffusion-lm improves controllable text generation.
\newblock In Sanmi Koyejo, S.~Mohamed, A.~Agarwal, Danielle Belgrave, K.~Cho, and A.~Oh, editors, \emph{Advances in Neural Information Processing Systems 35: Annual Conference on Neural Information Processing Systems 2022, NeurIPS 2022, New Orleans, LA, USA, November 28 - December 9, 2022}, 2022.

\bibitem[Liu et~al.(2025)Liu, Yang, Zhang, Chen, Zou, Wei, Wang, and Zhang]{DBLP:journals/corr/abs-2506-06295}
Zhiyuan Liu, Yicun Yang, Yaojie Zhang, Junjie Chen, Chang Zou, Qingyuan Wei, Shaobo Wang, and Linfeng Zhang.
\newblock dllm-cache: Accelerating diffusion large language models with adaptive caching.
\newblock \emph{CoRR}, abs/2506.06295, 2025.

\bibitem[Lu et~al.(2022)Lu, Mishra, Xia, Qiu, Chang, Zhu, Tafjord, Clark, and Kalyan]{DBLP:conf/nips/LuMX0CZTCK22}
Pan Lu, Swaroop Mishra, Tanglin Xia, Liang Qiu, Kai{-}Wei Chang, Song{-}Chun Zhu, Oyvind Tafjord, Peter Clark, and Ashwin Kalyan.
\newblock Learn to explain: Multimodal reasoning via thought chains for science question answering.
\newblock In \emph{NeurIPS}, 2022.

\bibitem[Lu et~al.(2024)Lu, Bansal, Xia, Liu, Li, Hajishirzi, Cheng, Chang, Galley, and Gao]{DBLP:conf/iclr/LuBX0LH0CG024}
Pan Lu, Hritik Bansal, Tony Xia, Jiacheng Liu, Chunyuan Li, Hannaneh Hajishirzi, Hao Cheng, Kai{-}Wei Chang, Michel Galley, and Jianfeng Gao.
\newblock Mathvista: Evaluating mathematical reasoning of foundation models in visual contexts.
\newblock In \emph{ICLR}, 2024.

\bibitem[Mei et~al.(2025)Mei, Yao, Ge, Wang, Bi, Cai, Liu, Li, Li, Zhang, Zhou, Mao, Xia, Guo, and Liu]{mei2025surveycontextengineeringlarge}
Lingrui Mei, Jiayu Yao, Yuyao Ge, Yiwei Wang, Baolong Bi, Yujun Cai, Jiazhi Liu, Mingyu Li, Zhong-Zhi Li, Duzhen Zhang, Chenlin Zhou, Jiayi Mao, Tianze Xia, Jiafeng Guo, and Shenghua Liu.
\newblock A survey of context engineering for large language models, 2025.
\newblock URL \url{https://arxiv.org/abs/2507.13334}.

\bibitem[Nichol et~al.(2022)Nichol, Dhariwal, Ramesh, Shyam, Mishkin, McGrew, Sutskever, and Chen]{DBLP:conf/icml/NicholDRSMMSC22}
Alexander~Quinn Nichol, Prafulla Dhariwal, Aditya Ramesh, Pranav Shyam, Pamela Mishkin, Bob McGrew, Ilya Sutskever, and Mark Chen.
\newblock {GLIDE:} towards photorealistic image generation and editing with text-guided diffusion models.
\newblock In Kamalika Chaudhuri, Stefanie Jegelka, Le~Song, Csaba Szepesv{\'{a}}ri, Gang Niu, and Sivan Sabato, editors, \emph{International Conference on Machine Learning, {ICML} 2022, 17-23 July 2022, Baltimore, Maryland, {USA}}, volume 162 of \emph{Proceedings of Machine Learning Research}, pages 16784--16804. {PMLR}, 2022.

\bibitem[Nie et~al.(2025)Nie, Zhu, You, Zhang, Ou, Hu, Zhou, Lin, Wen, and Li]{DBLP:journals/corr/abs-2502-09992}
Shen Nie, Fengqi Zhu, Zebin You, Xiaolu Zhang, Jingyang Ou, Jun Hu, Jun Zhou, Yankai Lin, Ji{-}Rong Wen, and Chongxuan Li.
\newblock Large language diffusion models.
\newblock \emph{CoRR}, abs/2502.09992, 2025.

\bibitem[OpenAI(2022)]{openai2022chatgpt}
OpenAI.
\newblock Chatgpt: Optimizing language models for dialogue.
\newblock OpenAI Blog, November 2022.
\newblock URL \url{https://openai.com/blog/chatgpt/}.

\bibitem[Ou et~al.(2025)Ou, Nie, Xue, Zhu, Sun, Li, and Li]{DBLP:conf/iclr/OuNXZSLL25}
Jingyang Ou, Shen Nie, Kaiwen Xue, Fengqi Zhu, Jiacheng Sun, Zhenguo Li, and Chongxuan Li.
\newblock Your absorbing discrete diffusion secretly models the conditional distributions of clean data.
\newblock In \emph{The Thirteenth International Conference on Learning Representations, {ICLR} 2025, Singapore, April 24-28, 2025}. OpenReview.net, 2025.

\bibitem[Radford et~al.(2018)Radford, Narasimhan, Salimans, Sutskever, et~al.]{radford2018improving}
Alec Radford, Karthik Narasimhan, Tim Salimans, Ilya Sutskever, et~al.
\newblock Improving language understanding by generative pre-training.
\newblock 2018.

\bibitem[Radford et~al.(2019)Radford, Wu, Child, Luan, Amodei, Sutskever, et~al.]{radford2019language}
Alec Radford, Jeffrey Wu, Rewon Child, David Luan, Dario Amodei, Ilya Sutskever, et~al.
\newblock Language models are unsupervised multitask learners.
\newblock \emph{OpenAI blog}, 1\penalty0 (8):\penalty0 9, 2019.

\bibitem[Rombach et~al.(2022)Rombach, Blattmann, Lorenz, Esser, and Ommer]{DBLP:conf/cvpr/RombachBLEO22}
Robin Rombach, Andreas Blattmann, Dominik Lorenz, Patrick Esser, and Bj{\"{o}}rn Ommer.
\newblock High-resolution image synthesis with latent diffusion models.
\newblock In \emph{{IEEE/CVF} Conference on Computer Vision and Pattern Recognition, {CVPR} 2022, New Orleans, LA, USA, June 18-24, 2022}, pages 10674--10685. {IEEE}, 2022.

\bibitem[Saharia et~al.(2022)Saharia, Chan, Saxena, Li, Whang, Denton, Ghasemipour, Lopes, Ayan, Salimans, Ho, Fleet, and Norouzi]{DBLP:conf/nips/SahariaCSLWDGLA22}
Chitwan Saharia, William Chan, Saurabh Saxena, Lala Li, Jay Whang, Emily~L. Denton, Seyed Kamyar~Seyed Ghasemipour, Raphael~Gontijo Lopes, Burcu~Karagol Ayan, Tim Salimans, Jonathan Ho, David~J. Fleet, and Mohammad Norouzi.
\newblock Photorealistic text-to-image diffusion models with deep language understanding.
\newblock In Sanmi Koyejo, S.~Mohamed, A.~Agarwal, Danielle Belgrave, K.~Cho, and A.~Oh, editors, \emph{Advances in Neural Information Processing Systems 35: Annual Conference on Neural Information Processing Systems 2022, NeurIPS 2022, New Orleans, LA, USA, November 28 - December 9, 2022}, 2022.

\bibitem[Sahoo et~al.(2024)Sahoo, Arriola, Schiff, Gokaslan, Marroquin, Chiu, Rush, and Kuleshov]{DBLP:conf/nips/SahooASGMCRK24}
Subham~S. Sahoo, Marianne Arriola, Yair Schiff, Aaron Gokaslan, Edgar Marroquin, Justin~T. Chiu, Alexander Rush, and Volodymyr Kuleshov.
\newblock Simple and effective masked diffusion language models.
\newblock In Amir Globersons, Lester Mackey, Danielle Belgrave, Angela Fan, Ulrich Paquet, Jakub~M. Tomczak, and Cheng Zhang, editors, \emph{Advances in Neural Information Processing Systems 38: Annual Conference on Neural Information Processing Systems 2024, NeurIPS 2024, Vancouver, BC, Canada, December 10 - 15, 2024}, 2024.

\bibitem[Seo et~al.(2017)Seo, Kembhavi, Farhadi, and Hajishirzi]{seo2016bidirectional}
Minjoon Seo, Aniruddha Kembhavi, Ali Farhadi, and Hannaneh Hajishirzi.
\newblock Bidirectional attention flow for machine comprehension.
\newblock In \emph{ICLR}, 2017.

\bibitem[Shi et~al.(2024)Shi, Han, Wang, Doucet, and Titsias]{DBLP:conf/nips/ShiHWDT24}
Jiaxin Shi, Kehang Han, Zhe Wang, Arnaud Doucet, and Michalis~K. Titsias.
\newblock Simplified and generalized masked diffusion for discrete data.
\newblock In Amir Globersons, Lester Mackey, Danielle Belgrave, Angela Fan, Ulrich Paquet, Jakub~M. Tomczak, and Cheng Zhang, editors, \emph{Advances in Neural Information Processing Systems 38: Annual Conference on Neural Information Processing Systems 2024, NeurIPS 2024, Vancouver, BC, Canada, December 10 - 15, 2024}, 2024.

\bibitem[Sohl{-}Dickstein et~al.(2015)Sohl{-}Dickstein, Weiss, Maheswaranathan, and Ganguli]{DBLP:journals/corr/Sohl-DicksteinW15}
Jascha Sohl{-}Dickstein, Eric~A. Weiss, Niru Maheswaranathan, and Surya Ganguli.
\newblock Deep unsupervised learning using nonequilibrium thermodynamics.
\newblock \emph{CoRR}, abs/1503.03585, 2015.

\bibitem[Song et~al.(2021)Song, Sohl{-}Dickstein, Kingma, Kumar, Ermon, and Poole]{DBLP:conf/iclr/0011SKKEP21}
Yang Song, Jascha Sohl{-}Dickstein, Diederik~P. Kingma, Abhishek Kumar, Stefano Ermon, and Ben Poole.
\newblock Score-based generative modeling through stochastic differential equations.
\newblock In \emph{9th International Conference on Learning Representations, {ICLR} 2021, Virtual Event, Austria, May 3-7, 2021}. OpenReview.net, 2021.

\bibitem[Stechly et~al.(2023)Stechly, Marquez, and Kambhampati]{DBLP:journals/corr/abs-2310-12397}
Kaya Stechly, Matthew Marquez, and Subbarao Kambhampati.
\newblock {GPT-4} doesn't know it's wrong: An analysis of iterative prompting for reasoning problems.
\newblock \emph{CoRR}, abs/2310.12397, 2023.

\bibitem[Valmeekam et~al.(2023)Valmeekam, Marquez, and Kambhampati]{DBLP:journals/corr/abs-2310-08118}
Karthik Valmeekam, Matthew Marquez, and Subbarao Kambhampati.
\newblock Can large language models really improve by self-critiquing their own plans?
\newblock \emph{CoRR}, abs/2310.08118, 2023.

\bibitem[Vedantam et~al.(2015)Vedantam, Zitnick, and Parikh]{DBLP:conf/cvpr/VedantamZP15}
Ramakrishna Vedantam, C.~Lawrence Zitnick, and Devi Parikh.
\newblock Cider: Consensus-based image description evaluation.
\newblock In \emph{CVPR}, pages 4566--4575, 2015.

\bibitem[Wang et~al.(2024)Wang, Pan, Shi, Lu, Ren, Zhou, Zhan, and Li]{DBLP:conf/nips/WangPSLRZZL24}
Ke~Wang, Junting Pan, Weikang Shi, Zimu Lu, Houxing Ren, Aojun Zhou, Mingjie Zhan, and Hongsheng Li.
\newblock Measuring multimodal mathematical reasoning with math-vision dataset.
\newblock In \emph{NeurIPS}, 2024.

\bibitem[Wu et~al.(2025)Wu, Zhang, Xue, Liu, Diao, Zhu, Luo, Han, and Xie]{DBLP:journals/corr/abs-2505-22618}
Chengyue Wu, Hao Zhang, Shuchen Xue, Zhijian Liu, Shizhe Diao, Ligeng Zhu, Ping Luo, Song Han, and Enze Xie.
\newblock Fast-dllm: Training-free acceleration of diffusion {LLM} by enabling {KV} cache and parallel decoding.
\newblock \emph{CoRR}, abs/2505.22618, 2025.

\bibitem[Yang et~al.(2025)Yang, Tian, Li, Zhang, Shen, Tong, and Wang]{DBLP:journals/corr/abs-2505-15809}
Ling Yang, Ye~Tian, Bowen Li, Xinchen Zhang, Ke~Shen, Yunhai Tong, and Mengdi Wang.
\newblock Mmada: Multimodal large diffusion language models.
\newblock \emph{CoRR}, abs/2505.15809, 2025.

\bibitem[Ye et~al.(2025)Ye, Xie, Zheng, Gao, Wu, Jiang, Li, and Kong]{dream2025}
Jiacheng Ye, Zhihui Xie, Lin Zheng, Jiahui Gao, Zirui Wu, Xin Jiang, Zhenguo Li, and Lingpeng Kong.
\newblock Dream 7b, 2025.
\newblock URL \url{https://hkunlp.github.io/blog/2025/dream}.

\bibitem[Young et~al.(2014)Young, Lai, Hodosh, and Hockenmaier]{DBLP:journals/tacl/YoungLHH14}
Peter Young, Alice Lai, Micah Hodosh, and Julia Hockenmaier.
\newblock From image descriptions to visual denotations: New similarity metrics for semantic inference over event descriptions.
\newblock \emph{Trans. Assoc. Comput. Linguistics}, 2:\penalty0 67--78, 2014.

\bibitem[Yue et~al.(2024)Yue, Ni, Zheng, Zhang, Liu, Zhang, Stevens, Jiang, Ren, Sun, Wei, Yu, Yuan, Sun, Yin, Zheng, Yang, Liu, Huang, Sun, Su, and Chen]{DBLP:conf/cvpr/YueNZ0LZSJRSWYY24}
Xiang Yue, Yuansheng Ni, Tianyu Zheng, Kai Zhang, Ruoqi Liu, Ge~Zhang, Samuel Stevens, Dongfu Jiang, Weiming Ren, Yuxuan Sun, Cong Wei, Botao Yu, Ruibin Yuan, Renliang Sun, Ming Yin, Boyuan Zheng, Zhenzhu Yang, Yibo Liu, Wenhao Huang, Huan Sun, Yu~Su, and Wenhu Chen.
\newblock {MMMU:} {A} massive multi-discipline multimodal understanding and reasoning benchmark for expert {AGI}.
\newblock In \emph{CVPR}, pages 9556--9567. {IEEE}, 2024.

\bibitem[Zhao et~al.(2025)Zhao, Gupta, Zheng, and Grover]{DBLP:journals/corr/abs-2504-12216}
Siyan Zhao, Devaansh Gupta, Qinqing Zheng, and Aditya Grover.
\newblock d1: Scaling reasoning in diffusion large language models via reinforcement learning.
\newblock \emph{CoRR}, abs/2504.12216, 2025.

\end{thebibliography}
\bibliographystyle{plainnat}    
}
\appendix
\clearpage

\section{Case Study}\label{appendix: case}
\textbf{Case Study: Decoding Dynamics.} To conduct a fine-grained examination of the decoding processes of the baselines and {\method}, we present two generation examples---one from GSM8K and one from MMMU-val below. As shown, the baseline may produce erroneous tokens at the early decoding stages. Since the generated tokens by the baseline remain unchanged in subsequent decoding steps, the false contextual information propagates throughout the whole generation process, eventually leading to low-quality generation results. In contrast, {\method} enables dynamical refinement of generated tokens via an iterative draft-and-verify mechanism, which mitigates error accumulation and facilitates high-quality decoded outputs.

\begin{tcolorbox}[title = {Case: GSM8K, Geneation Length 256, Block Length 128, 2 blocks in total}, breakable, fontupper=\small, fonttitle=\small]

\textbf{Question: }In a family, there are 2 brothers and 3 sisters.  All sisters are the same age, which is 16.  One of the brothers is 12 years old,  which is half the age of the older brother.  What is the total age of all these siblings?

\textbf{GT: }84.

\noindent\rule{\linewidth}{0.5pt}

\textbf{LLaDA: }

\noindent\hdashrule{\linewidth}{0.5pt}{2pt 2pt}

{\scriptsize\bf
\color{gray} Step: 128 (\textcolor{red}{1st block decoding completed}). 1st block:}

{\scriptsize\ttfamily
<reasoning>

1. We know that all three sisters are 16 years old.

2. One of the brothers is 12 years old, which is half the age of the older brother.

3. Let's denote the age of the older brother as \( x \).

4. According to the problem, 12 is half the age of the older brother, so we can write the equation: \( 12 = \frac{x}{2} \).

5. Solving for \( x \), we multiply both sides by 2: \( x = 24 \).

6. Therefore, the
}

\noindent\hdashrule{\linewidth}{0.5pt}{2pt 2pt}

{\scriptsize\bf
\color{gray} Step: 162 (\textcolor{red}{early overconfident error}). 2nd block:}

{\scriptsize\ttfamily
older brother is 24 years old.

7.<|mdm\_mask|>……<|mdm\_mask|>

</reasoning>

<answer>

\boxed{\textcolor{red}{96}}

</answer><|eot\_id|><|endoftext|>}

\noindent\hdashrule{\linewidth}{0.5pt}{2pt 2pt}

{\scriptsize\bf
\color{gray} Final (\textcolor{red}{error accumulation}):}

{\scriptsize\ttfamily
<reasoning>

1. We know that all three sisters are 16 years old.

2. One of the brothers is 12 years old, which is half the age of the older brother.

3. Let's denote the age of the older brother as \( x \).

4. According to the problem, 12 is half the age of the older brother, so we can write the equation: \( 12 = \frac{x}{2} \).

5. Solving for \( x \), we multiply both sides by 2: \( x = 24 \).

6. Therefore, the older brother is 24 years old.

7. The younger brother, being the only other brother, \textcolor{red}{is also 24 years old}.

8. The total age of the two brothers is {\color{red}\( 24 + 24 = 48 \)}.

9. The total age of the three sisters is \( 3 \times 16 = 48 \).

10. The total age of all the siblings is {\color{red}\( 48 + 48 = 96 \)}.

</reasoning>

<answer>

\boxed{\textcolor{red}{96}}

</answer><|eot\_id|><|endoftext|>
}

\noindent\rule{\linewidth}{0.5pt}

\textbf{{\method}: }

\noindent\hdashrule{\linewidth}{0.5pt}{2pt 2pt}

{\scriptsize\bf
\color{gray} Step: 35 (\textcolor{red}{1st block decoding completed}). 1st block:}

{\scriptsize\ttfamily
<reasoning>

1. We know that all three sisters are 16 years old.

2. One of the brothers is 12 years old, which is half the age of the older brother.

3. Let's denote the age of the older brother as \( x \).

4. According to the problem, 12 is half the age of the older brother, so we can write the equation: \( 12 = \frac{x}{2} \).

5. Solving for \( x \), we multiply both sides by 2: \( x = 24 \).

6. Therefore, the
}

\noindent\hdashrule{\linewidth}{0.5pt}{2pt 2pt}

{\scriptsize\bf
\color{gray} Step: 37 (\textcolor{red}{draft error}). 2nd block:}

{\scriptsize\ttfamily
older brother is 24 years old.

7.<|mdm\_mask|>……<|mdm\_mask|> = 96<|mdm\_mask|><|mdm\_mask|><|mdm\_mask|>

</reasoning<|mdm\_mask|>

<answer>

\boxed{\textcolor{red}{9}\text{<|mdm\_mask|>}}

</answer><|eot\_id|><|endoftext|>
}

\noindent\hdashrule{\linewidth}{0.5pt}{2pt 2pt}

{\scriptsize\bf
\color{gray} Step: 42 (\textcolor{red}{refinement: still wrong}). 2nd block:}

{\scriptsize\ttfamily
 older brother is 24 years old.
 
7. The total age of<|mdm\_mask|> sisters<|mdm\_mask|> \textbackslash( 3<|mdm\_mask|>times 16 = 4<|mdm\_mask|>……<|mdm\_mask|> total age of all the<|mdm\_mask|> is \textbackslash( 48 + <|mdm\_mask|><|mdm\_mask|> = \textcolor{red}{86} \textbackslash) years.

</reasoning>

<answer>

\boxed{\textcolor{red}{86}}

</answer><|eot\_id|><|endoftext|>

}

\noindent\hdashrule{\linewidth}{0.5pt}{2pt 2pt}

{\scriptsize\bf
\color{gray} Final (\textcolor{red}{all correct through iterative refinement}):}

{\scriptsize\ttfamily
<reasoning>

1. We know that all three sisters are 16 years old.

2. One of the brothers is 12 years old, which is half the age of the older brother.

3. Let's denote the age of the older brother as \( x \).

4. According to the problem, 12 is half the age of the older brother, so we can write the equation: \( 12 = \frac{x}{2} \).

5. Solving for \( x \), we multiply both sides by 2: \( x = 24 \).

6. Therefore, the older brother is 24 years old.

7. The total age of the sisters is \( 3 \times 16 = 48 \) years.

8. The total age of the brothers is the sum of the ages of the 12-year-old and the 24-year-old brother: \( 12 + 24 = 36 \) years.

9. The total age of all the siblings is \( 48 + 36 = 84 \) years.

</reasoning>

<answer>

\boxed{\textcolor{red}{84}}

</answer><|eot\_id|><|endoftext|>
}

\end{tcolorbox}

\begin{tcolorbox}[title = {Case: MMMU-val, Geneation Length 256, Block Length 128, 2 blocks in total}, breakable, fontupper=\small, fonttitle=\small]

\vspace{\medskipamount}
{\includegraphics[width=0.5\textwidth]{Fig/mmmu_demo.png}}
\vspace{\medskipamount}

\textbf{Question: } The graph of \( y = f(x) \) is shown in the figure above. The shaded region \( A \) has area \( a \) and the shaded region \( B \) has area \( b \). If \( g(x) = f(x) +3 \), what is the average value of g on the interval \( [-2,4] \)? \\
A. \( (a+b+3)/6 \) \\
B. \( (-a+b+3)/6 \) \\
C. \( (-a+b)/6+3 \) \\
D. \( (a+b)/6+3 \) \\

\textbf{GT: }C.

\noindent\rule{\linewidth}{0.5pt}

\textbf{MMaDA: }

\noindent\hdashrule{\linewidth}{0.5pt}{2pt 2pt}

{\scriptsize\bf
\color{gray} Step: 70 (\textcolor{red}{early overconfident error}). 1st block:}

{\scriptsize\ttfamily
<think>

1. To find the average value of \( g(x) \) on the interval \([-2, 4]\), we need to calculate the area of the graph of \( g(x) \) over this interval. 

2. \textcolor{red}{The area of the graph of \( g(x) \) is given by the sum of the areas of the two shaded regions.}

<|mdm\_mask|>……<|mdm\_mask|>
}

\noindent\hdashrule{\linewidth}{0.5pt}{2pt 2pt}

{\scriptsize\bf
\color{gray} Step: 128 (\textcolor{red}{error accumulation}). 1st block:}

{\scriptsize\ttfamily
<think>

1. To find the average value of \( g(x) \) on the interval \([-2, 4]\), we need to calculate the area of the graph of \( g(x) \) over this interval. 

2. \textcolor{red}{The area of the graph of \( g(x) \) is given by the sum of the areas of the two shaded regions.} 

3. The area of region \( A \) is \( a \) and the area of region \( B \) is \( b \). 

4. \textcolor{red}{Therefore, the total area of \( g(x) \) is \( a + b \).}

5. The average value of \( g(x) \) the interval
}

\noindent\hdashrule{\linewidth}{0.5pt}{2pt 2pt}

{\scriptsize\bf
\color{gray} Final (\textcolor{red}{error accumulation}): }

{\scriptsize\ttfamily
<think>

1. To find the average value of \( g(x) \) on the interval \([-2, 4]\), we need to calculate the area of the graph of \( g(x) \) over this interval. 

2. \textcolor{red}{The area of the graph of \( g(x) \) is given by the sum of the areas of the two shaded regions.}

3. The area of region \( A \) is \( a \) and the area of region \( B \) is \( b \). 

4. \textcolor{red}{Therefore, the total area of \( g(x) \) is \( a + b \).}

5. The average value of \( g(x) \) on the interval \([-2, 4]\) is \textcolor{red}{\( \frac{a+b}{6} \).}

</think>

\textcolor{red}{A}<|endoftext|>
}

\noindent\rule{\linewidth}{0.5pt}

\textbf{{\method}: }

\noindent\hdashrule{\linewidth}{0.5pt}{2pt 2pt}

{\scriptsize\bf
\color{gray} Step: 13 (\textcolor{red}{draft error}). 1st block:}

{\scriptsize\ttfamily
<think>

1. To find the average value of \( g(x) \) on the interval \([-2, 4]\), we need to use the<|mdm\_mask|> for the<|mdm\_mask|> of \( f(x) \) over<|mdm\_mask|> interval. 

2. The average value of <|mdm\_mask|>\((x) \) is given by \textcolor{red}{\textbackslash frac\{1\}\{2<|mdm\_mask|> \textbackslash times \textbackslash<|mdm\_mask|><|mdm\_mask|>2\}\textasciicircum\{4}<|mdm\_mask|>……<|mdm\_mask|>
}

\noindent\hdashrule{\linewidth}{0.5pt}{2pt 2pt}

{\scriptsize\bf
\color{gray} Step: 33 (\textcolor{red}{refinement: correct}). 1st block:}

{\scriptsize\ttfamily
<think>

1. To find the average value of \( g(x) \) on the interval \([-2, 4]\), we need to use the formula for the average of \( f(x) \) over an interval. 

2. The average value of \( f(x) \) is given by \textcolor{red}{\(\frac{1}{6} \cdot \int_{-2}^{4} f(x) \, dx\).}

3. Given that \( g(x) = f(x) + 3 \), we need to find \( f(x) \). The average value of \( f(x) \) is \(\frac{-a+b}{6}\). 

4. Therefore, <|mdm\_mask|>……<|mdm\_mask|>

}

\noindent\hdashrule{\linewidth}{0.5pt}{2pt 2pt}

{\scriptsize\bf
\color{gray} Final (\textcolor{red}{all correct through iterative refinement}):}

{\scriptsize\ttfamily
<think>

1. To find the average value of \( g(x) \) on the interval \([-2, 4]\), we need to use the formula for the average of \( f(x) \) over an interval. 

2. The average value of \( f(x) \) is given by \(\frac{1}{6} \cdot \int_{-2}^{4} f(x) \, dx\).

3. Given that \( g(x) = f(x) + 3 \), we need to find \( f(x) \). The average value of \( f(x) \) is \(\frac{-a+b}{6}\). 

4. Therefore, the average value of \( g(x) \) on the interval \([-2, 4]\) is \(\frac{-a+b}{6} + 3\).

5. The correct answer is option \textcolor{red}{C}.

</think>

\textcolor{red}{C}<|endoftext|>
}

\end{tcolorbox}

\section{Limitations and Future Directions}
Our work establishes {\method} as a training-free framework that makes the decoding process of Diffusion Large Language Models (DLLMs) revokable, effectively addressing their quality-speed trade-off. The degree of acceleration, however, is inherently linked to the base model's capabilities; as our experiments show, more proficient models produce better drafts that require fewer refinement steps, leading to greater speedups. This insight points to a promising future direction: integrating the concept of revokable sampling directly into the training phase. A model trained with an awareness of a draft-and-verify mechanism could learn to generate more robust initial predictions and self-correct more efficiently, potentially unlocking even greater gains in performance and speed.

\end{document}